\documentclass[%
 reprint,
 superscriptaddress,
 amsmath,amssymb,
 aps,
 linenumbers,
]{revtex4-2}

\usepackage{preamble}

\newcommand{\pasteuraffil}{Institut Pasteur, Universit\'e Paris Cit\'e, CNRS UMR 3571, Decision and Bayesian Computation, 75015 Paris, France.}
\newcommand{\inriaaffil}{\'{E}pimeth\'ee, Inria, Paris, France.}

\begin{document}

\preprint{APS/123-QED}
\nolinenumbers
\title{Information maximization for a broad variety of multi-armed bandit games}

\author{Alex Barbier--Chebbah}
\email{alex.barbier-chebbah@pasteur.fr}%
\affiliation{\pasteuraffil}
\affiliation{\inriaaffil}

\author{Christian L.\ Vestergaard}
\affiliation{\pasteuraffil}
\affiliation{\inriaaffil}

\author{Jean-Baptiste Masson}
\email{jbmasson@pasteur.fr}%
\affiliation{\pasteuraffil}
\affiliation{\inriaaffil}

\date{\today}


\begin{abstract}
Information and free-energy maximization are physics principles that provide general rules for an agent to optimize actions in line with specific goals and policies. These principles are the building blocks for designing decision-making policies capable of efficient performance with only partial information. Notably, the information maximization principle has shown remarkable success in the classical bandit problem and has recently been shown to yield optimal algorithms for Gaussian and sub-Gaussian reward distributions. This article explores a broad extension of physics-based approaches to more complex and structured bandit problems. 
To this end, we cover three distinct types of bandit problems, where information maximization is adapted and leads to strong performances. Since the main challenge of information maximization lies in avoiding over-exploration, we highlight how information is tailored at various levels to mitigate this issue, paving the way for more efficient and robust decision-making strategies.
\end{abstract}

\maketitle

\section{Introduction}

\begin{figure*}[ht]
\centering
\includegraphics[scale=0.85]{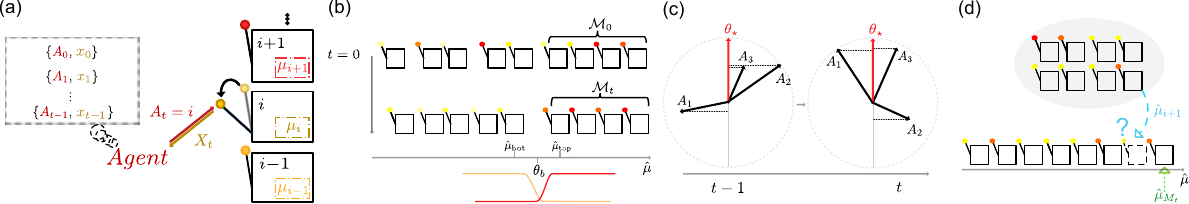}
\caption{Illustrations of multi-armed bandit  settings addressed in this work. 
\textbf{a)} Illustration of the multi-armed bandit principle. At each time step $t$ , the agent chooses an action $i=A_t$ that returns a reward $x_t$ drawn from a distribution of unknown mean $\mu_{i}$. The arm is denoted as $A_t$ when it is a vector and as $\armt$ when it is an index.
By accumulating rewards, the agent also gathers information about the average rewards of the arms and optimizes its decision-making policy. The specific goals of the agent will vary depending on the bandit problem. \textbf{b)} General principle of the Explore-$m$ problem. Here, the goal is to identify the $m$ arms with the highest mean rewards (red arms). To this end, our approach introduces a well-designed separator $\meanb$ between the current $m$ highest arms (forming the set $\mathcal{M}_t$) and the remaining arms at each time. Our algorithm chooses the arm to maximize the information gain of the separator's effectiveness in distinguishing both subsets (cumulative distribution pictured in yellow and red below). \textbf{c)} Illustration of the linear bandit setting. Here, arms are d-dimensional vectors resampled at each time step. The scalar between the arm vector and a constant but unknown vector gives the mean reward value $\Astar$. Because of this geometric dependency on the reward, pulling an arm also provides information on the expected results of the others. \textbf{d)} Illustration of the may-arms settings. Here, the agent has a finite duration, known in advance to maximizes its gains. Because there is no time to explore all the arms thoroughly, the agent needs to focus on a small subset of arms to isolate a sufficiently promising solution before time runs out. Our approach balances two possible actions: exploring a new arm (represented in blue) or greedily exploiting the current empirical best solution (represented in green) based on the empirical mean $\mu_{M_{t}}$. We model the upcoming losses as a function to guide the decision-making process.}
\label{fig:schema}
\end{figure*} 

Decision-making under partial information is a critical challenge faced by biological organisms (in particular relating to taxis or cognition \cite{suttonReinforcementLearningIntroduction1998a,goldNeuralBasisDecision2007,jepmaPupilDiameterPredicts2011,kimRoleStriatumUpdating2009}), but also extends to broader fields such as Bayesian optimization, robotics and machine learning \citep{vergassolaInfotaxisStrategySearching2007, murlisOdorPlumesHow1992,massonChasingInformationSearch2009b, wangMaxvalueEntropySearch2017, russoLearningOptimizeInformationDirected2014, lindnerInformationDirectedReward2022, houthooftVIMEVariationalInformation2017,dingInteractiveAnomalyDetection2019}. 
To perform effectively, the decision-making agent must account for the physical structure of both the environment and decision problem to formulate its strategy, which often involves incorporating complex priors or tailored design elements \cite{zhangMultiRobotSearchingSparse2015c, hernandez-lobatoPredictiveEntropySearch2016, martinezUsingInsectElectroantennogram2014}. To this end, general physics principles can serve as key tools, offering observables that inherently adapt to the underlying structure of the environment \cite{cardeNavigationWindbornePlumes2021, cohenShouldStayShould2007, doyaBayesianBrainProbabilistic2007, hillsExplorationExploitationSpace2015a}. These principles are also designed to adapt to new information, even conflicting, or when the agent is limited to partial or noisy feedback from a fluctuating environment \cite{vergassolaInfotaxisStrategySearching2007}. They have been applied across various domains, including modeling decision-making in the brain using the free-energy principle \cite{fristonActiveInferenceLearning2016}, optimizing the accuracy-complexity trade-off through the information bottleneck principle \citep{Tishby2015}, and facilitating navigation in fluctuating environments via information maximization strategies \citep{vergassolaInfotaxisStrategySearching2007}.

In the information-maximization framework, the agent seeks to maximize information about one or more relevant observables. The agent's choice is thus driven by the option that maximizes the expected information gain. Due to its versatility in accommodating various environmental structures, this principle has proven its broad applicability across both applied and theoretical areas \cite{heliasStatisticalFieldTheory2020a, parrActiveInferenceFree2022}. Notable applications include robotics, where information sharing enhances collective decision-making \cite{zhangMultiRobotSearchingSparse2015c}, and taxis with the search for olfactory sources in turbulent flows \cite{massonOlfactorySearchesLimited2013b, reddyOlfactorySensingNavigation2022}. Maximum entropy principles also underpin approaches in Bayesian optimization and reinforcement learning \cite{lindnerInformationDirectedReward2022, houthooftVIMEVariationalInformation2017,hernandez-lobatoPredictiveEntropySearch2016}.

Multi-armed bandit (MAB) problems constitute a generalized framework for addressing decision-making in fluctuating and partially known environments \cite{lattimoreBanditAlgorithms2020, suttonReinforcementLearningIntroduction1998a}, and has garnered significant attention in a wide range of applications, spanning from neuroscience and healthcare to recommender systems and finance \cite{lattimoreBanditAlgorithms2020,silverMasteringGameGo2016,bouneffoufSurveyApplicationsMultiArmed2020, burtiniSurveyOnlineExperiment2015, durandContextualBanditsAdapting2018, gentileOnlineClusteringBandits2014, kuleshovAlgorithmsMultiarmedBandit2014, maryBanditsRecommenderSystems2015, stewartLearningSelectActions2012, jepmaPupilDiameterPredicts2011}
In the multi-armed bandit problem, an agent is presented with a set of possible actions, or "arms", each associated with a probabilistic reward (akin to a multi-armed slot machine game). The agent must optimize its actions to achieve a specified goal, which generally involves maximizing cumulative rewards over a fixed or infinite horizon. However, it may occasionally involve identifying a subset of optimal arms or exploring its available options [\cref{fig:schema} \textbf{a)}]. Hence, classic bandit settings address the exploration-exploitation dilemma and provide strategies that may achieve asymptotic optimality \cite{laiAsymptoticallyEfficientAdaptive1985, kaufmannThompsonSamplingAsymptotically2012, auerUsingUpperConfidence2000}. MAB problems have rapidly expanded to cover many more complex decision scenarios. A first main class of problems focuses on pure exploration, where the agent's goal is solely to identify a single optimal action or a subset of promising actions, without incurring penalties for sampling suboptimal ones \cite{bubeckPureExplorationFinitelyarmed2011a, jamiesonLilUCBOptimal2014, kaufmannInformationComplexityBandit2013, kalyanakrishnanPACSubsetSelection2012a,degennePureExplorationMultiple2019} [\cref{fig:schema} \textbf{b)}]. In essence, pure exploration aims to optimize a test phase to identify reliable candidates using a minimal amount of resources while suffering no penalty for exploration. Following such a test phase, the exploitation phase may ensue, but it is not optimized in this context. Such challenges arise in lower-risk scenarios like cosmetic drug trials or numerical resource allocation. A complementary area is termed contextual bandits, which have found application for example in personalized recommendations and pandemic control \citep{liContextualbanditApproachPersonalized2010, linOptimalEpidemicControl2022}. A notable example is linear bandit problems where actions' outcomes, because of an underlying structure between arms, are correlated \cite{zhouSurveyContextualMultiarmed2016,lattimoreBanditAlgorithms2020,auerUsingConfidenceBounds2002, lattimoreEndOptimismAsymptotic2017, agrawalThompsonSamplingContextual2013}, allowing one to gain information about other actions than the one performed [\cref{fig:schema} \textbf{c)}].
Lastly, many-armed problems address scenarios where the number of available actions is large compared to the time horizon, requiring the agent to concentrate on sufficiently promising solutions [\cref{fig:schema} \textbf{d)}]. This challenge of finite resources resonates with applications in pharmaceutical and medical trials \cite{villarMultiarmedBanditModels2015, villarBanditsStrategiesEvaluated2018}. Such settings offer various extensions and readily apply to cases where the number of arms is infinite \cite{deheideBanditsManyOptimal2024,bonaldTwoTargetAlgorithmsInfiniteArmed2013, berryBanditProblemsInfinitely1997, wangAlgorithmsInfinitelyManyArmed2008}. Additional extensions of MAB problems explore dynamic rewards, adversarial settings, and multi-agent interactions, focusing on competition, congestion, and overexploitation respectively, but are not investigated further here \cite{lattimoreBanditAlgorithms2020,zhouSurveyContextualMultiarmed2016,degennePureExplorationMultiple2019,baudrySubsamplingEfficientNonparametric2020,guhaApproximationAlgorithmsRestless2010,levineRottingBandits2017}.

We address here three classes of active bandit problems to show how the general principles of information maximization can be adapted to address structured decision-making problems. Although a wide variety of bandit games and associated strategies have been studied in recent years, strategies developed to address them primarily rely on a few core principles, most notably confidence bounds on the individual arms' mean rewards, which frequently drive the decision-making process \cite{auerUsingUpperConfidence2000,auerFinitetimeAnalysisMultiarmed2002,bastaniMostlyExplorationFreeAlgorithms2021, menardMinimaxAsymptoticallyOptimal2017}. 
Although there are exceptions to this rule and several alternative approaches have been investigated \cite{kaufmannThompsonSamplingAsymptotically2012, agrawalThompsonSamplingContextual2013,russoInformationTheoreticAnalysisThompson2016, reddyInfomaxStrategiesOptimal2016, baudrySubsamplingEfficientNonparametric2020,hungRewardBiasedMaximumLikelihood2020,tirinzoniAsymptoticallyOptimalPrimaldual2020}, there is still a crucial need to develop original approaches, especially when addressing more complex bandit challenges.

 
Recently, the information maximization principle has been shown to achieve strong results in classical bandit settings, displaying state-of-the-art empirical performance both at short and long time horizons and when handling multiple arms. \citep{reddyInfomaxStrategiesOptimal2016, barbier-chebbahApproximateInformationEfficient2023}. 
Information maximization algorithms  (and information-based algorithms, more generally) aim to minimize the expected entropy (a functional) of a key variable within the system.

The first difficulty in applying information-based algorithms is that they rely on integrating such a function that often has no known analytical solution. Their application thus relies on numerical integration, which both hinders theoretical analysis of their statistical performance and increases their computational costs \citep{barbier-chebbahApproximateInformationEfficient2023}.
Recently, we introduced a new class of approximate information maximization (AIM) algorithms \citep{barbier-chebbahApproximateInformationEfficient2023, barbier-chebbahApproximateInformationMaximization2023} that rely on tractable approximations of the entropy. 
The analytical tractability of the AIM algorithms significantly increases their computational efficiency, opening up for their application to a wider range of bandit problems~\citep{barbier-chebbahApproximateInformationEfficient2023,barbier-chebbahApproximateInformationMaximization2023}. Furthermore, it enabled us to rigorously prove their asymptotic optimality in the classic setting with Gaussian and sub-Gaussian rewards~\citep{barbier-chebbahApproximateInformationMaximization2023}, confirming the heuristic observation that information maximization can drive optimal learning and optimization.

A second challenge with information-based approaches remains choosing the observable and functional to appropriately capture the underlying structure of the bandit problem and ensure an efficient policy. 
The main challenge of information maximization is to avoid over-exploration~\cite{reddyInfomaxStrategiesOptimal2016}, requiring careful design of the observables or features on which entropy minimization is performed as well as of the approximations performed to make the algorithms tractable~\cite{barbier-chebbahApproximateInformationEfficient2023}. 
In this work, we show how the approximate information maximization principle can be applied to the three distinct bandit topics discussed above, namely the pure exploration problem of explore-$m$ selection, contextual bandits exemplified by linear bandits, and many-armed bandits. 
Since the three problems each involve distinct levels of exploration, we  discuss how relevant information is accessed within each context. This allows us to extend our building principle to a broader range of structures not specifically meant for pure exploration.
We provide numerical evaluations of the performance of the approximate information-based approaches and compare it to well-established algorithms in each case. A key challenge left for future work is to establish theoretical bounds for the algorithms investigated below.

\section{Structure of this paper}

We detail our approach to the three distinct bandit settings mentioned above. For each case, we follow the same three steps: 
1) we introduce the bandit game, defining the key elements that are required of an efficient strategy, and provide an overview of existing strategies to tackle the problem; 
2) we detail our design principles for an information-based approach, followed by the development of our approximation scheme that yields a tractable expression and enables an efficient implementation; 
3) finally, we investigate the performance of the resulting algorithm, its specific features, and we discuss potential extensions.

\section{Explore-m bandits}

\subsection{Setting and core insights}

We first consider a well-known multiple-arm identification problem, commonly referred to as the Explore-$m$ problem under the probably approximately correct (PAC)-identification formulation \cite{kalyanakrishnanPACSubsetSelection2012a, kaufmannComplexityBestarmIdentification2016, kalyanakrishnanEfficientSelectionMultiple2010,  mannorSampleComplexityExploration2004, zhouOptimalPACMultiple2014}. In these problems, we aim to identify a subset of $\tk$ promising arms (i.e individual actions or choices) among $\totk$ possible options [\cref{fig:schema} \textbf{b)}]. Each arm is associated with a reward distribution of unknown mean parameter $\mean{i}$ (supported on $ [0,1] $ without loss of generality).  
At each time step $t$, the agent selects an arm $i_t$ and obtains an associated stochastic reward $x_t$ sampled from the $i$th arm's reward distribution.  By sampling the environment, the agent sequentially accumulates information on the parameters of the sampled arms' means $\mean{i}$. For simplicity, we sort the arms according to their (unknown) means $\mean{\totk}\leq \mean{n-1} <.. <\mean{m} <.. < \mean{1}$. 

Explore-$m$ aims to identify the subset of $\tk$ arms among the highest means up to $\epsilon$ termed $\epsilon$-optimal. More precisely, an arm $i$ is defined to be $\epsilon$-optimal, iff $\mean{i} \geq \mean{m} - \epsilon$ for some $\epsilon \in (0,1)$ where $\mean{m}$ is the $m$th-highest mean. This criterion prevents the algorithm from spending excessive time distinguishing between arms when $\mean{m}$ and $\mean{m+1}$ are infinitesimally close, ensuring it focuses on more meaningful scenarios. 

Since no additional cost is incurred by sampling suboptimal arms, the algorithm's efficiency is evaluated based solely on its ability to identify such a subset both quickly and with high confidence. Two standard evaluation criteria are commonly considered \cite{kaufmannInformationComplexityBandit2013}. 
In the fixed-budget setting, the algorithm has a finite amount of time to identify the subset, and its performance is measured through the probability of returning an $\epsilon$-optimal subset. 
In contrast, the PAC formulation requires the algorithm to stop when the probability of returning an $\epsilon$-optimal subset exceeds $1 - \delta$, with $\delta \in (0,1)$, with the goal of minimizing the stopping time \cite{kalyanakrishnanPACSubsetSelection2012a}. We focus here on the PAC formulation in the following. 

Algorithms for pure exploration problems can broadly be divided into two categories. In the \textbf{sample and reject strategies}, a subset of remaining arms is all sampled at each round. Next, a rejection criteria discards from the remaining subset sufficiently drawn arms. This process is repeated sequentially until a stopping criteria is met \cite{wangAlgorithmsInfinitelyManyArmed2008, bubeckMultipleIdentificationsMultiArmed2013}. For \textbf{adaptive sampling strategies}, at each time step, one or two arms with the most critical confidence regarding the subset identification are selected and pulled, with the arms' indices changing over time \cite{kalyanakrishnanPACSubsetSelection2012a,kaufmannInformationComplexityBandit2013,gabillonBestArmIdentification2012}. Most methods rely on upper and lower bounds on confidence intervals on the arms' means, rooted in the general class of upper confidence bound (UCB) algorithms \cite{jamiesonLilUCBOptimal2014,audibertBestArmIdentification2010, bubeckMultipleIdentificationsMultiArmed2013, gabillonBestArmIdentification2012}. By comparing the upper confidence bounds of likely suboptimal arms with the lower confidence bounds of the promising subset, these methods can derive an efficient decision-making procedure. However, the stopping criteria are highly dependent on the confidence interval design, which may become less effective when facing various reward distributions or more complex scenarios. For a deeper exploration of the sampling complexity of such strategies under different reward distributions, we refer to \cite{kaufmannComplexityBestarmIdentification2016, mannorSampleComplexityExploration2004}.

Since no explicit cost is incurred by selecting arms, we would expect a strategy focused solely on acquiring information about the best subset to be efficient. To this end, we developed a fully information-based strategy, ensuring feasibility and tractability through successive derivations, approximations, and simplifications. We focus on the PAC setting but discuss its extension to other pure exploration settings.

\subsection{Design principle}

We first introduce additional notation specific to the Explore-$m$ problem. 
In each round $t$, each arm $i$ is characterized by its empirical mean $\meanh{i} = \sum_{t : i_t=i} x_{t}/t$ and the number of times it has been drawn $\loct{i}$ (where we have omitted the subscript $t$ to avoid clutter). 
$\Ht = \{(i_1,x_1), (i_2,x_2), \ldots, (i_t,x_t)\}$ represents the complete history of chosen arms and observed rewards. 
The subset of arms with the current $\tk$ highest empirical means is denoted $\topset$ with associated sets of true (hidden) mean values, $\topsetmean$, and empirical means, $\topsetmeanh$. The equivalent notations for the remaining arms are $ \botset = \{ \totk \} \setminus \topset$, $ \botsetmean$, and $\botsetmeanh$, respectively. 
We also denote the worst selected mean by $\topmin = \min (\topsetmeanh)$ and the best unselected by $\botmax = \max(\botsetmeanh)$.

A promising candidate distribution for information maximization is the probability of $\topset$ being $\epsilon$-optimal, i.e., $ \Prob( \min ( \topsetmean) > \max(\botsetmean)-\epsilon | \Ht )$. Unfortunately, this posterior distribution is analytically intractable in practice. Additionally, since the arms are not identically distributed, applying asymptotic extreme value statistics is also challenging without approximations. 

Therefore, we adjust the decision process as follows. We introduce $\meanb$ as a separator value between the observed sets $\topsetmeanh$ and $\botsetmeanh$ and compare all the arms' mean likelihoods relative to $\meanb$. In other words, we now focus on comparing the arms' posterior probabilities of their means being above or below $\meanb$, given by
\begin{equation}\label{explorePtop}
\begin{split}
\Ptopset &= \Prob \left( \min ( \topsetmean) > \meanb-\frac{\epsilon}{2} | \Ht  \right)\\
&= \prod_{i \in \topset} \Prob \left( \mu_i \geq \meanb - \frac{\epsilon}{2} | \Ht \right),
\end{split}
\end{equation}
and 
\begin{equation}\label{explorePbot}
\begin{split}
\Pbotset &= \Prob \left( \max (\botsetmean) < \meanb +\frac{\epsilon}{2} | \Ht \right) \\
&=\prod_{i \in \botset} \Prob \left( \mu_i \leq \meanb + \frac{\epsilon}{2} | \Ht \right) ,
\end{split}
\end{equation}
respectively.

Notably, if both events occur, $\topset$ is $\epsilon$-optimal, making \cref{explorePtop,explorePbot} an effective proxy to derive an information maximization strategy. 
This leads us to a policy that selects the arm maximizing $\Pbotset \Ptopset$ growth. Leveraging the arms' independence, we can derive a straightforward expression for their expected increases should each arm be selected (see Supplementary Material \cref{Supp:SecExplore} for details):
\begin{equation}
\frac{\Delta_{i \in \topset } \Ptopset \Pbotset}{\Ptopset \Pbotset} = \frac{ \Delta \Prob \left( \mu_i \geq \meanb - \frac{\epsilon}{2} | \Ht \right)}{\Prob \left( \mu_i \geq \meanb - \frac{\epsilon}{2} | \Ht \right)},
\end{equation}
and 

\begin{equation}
\frac{\Delta_{i \in \botset } \Pbotset \Ptopset}{\Ptopset \Pbotset} = \frac{\Delta \Prob \left( \mu_i \leq \meanb + \frac{\epsilon}{2} | \Ht \right) }{\Prob \left( \mu_i \leq \meanb + \frac{\epsilon}{2} | \Ht \right)}.
\end{equation}

where the increment $\Delta$ symbols may either account for the expected discrete increment or continuous gradient along arm $i$. Exact expressions can be derived for both choices, and for simplicity, we opt for the latter one (see Supplementary Material \cref{suppSecPexp}). The selected arm is then given by
\begin{equation}\label{explorem_armselect}
 \armt = \underset{i \in \topset \cup \botset}{\argmax} \left( \Delta_{i \in \botset } \Pbotset \Pbotset, \Delta_{i \in \topset } \Ptopset \Pbotset \right) .
\end{equation}

This approach offers two main advantages. First, the information gain is significantly simplified as it relies on a single-point evaluation. Second, it directly provides a well-defined stopping criterion for the PAC setting. As noted above, the joint probabilities \cref{explorePtop,explorePbot} lower bounds the probability of returning an $\epsilon$-optimal subset. As both $\Ptopset$ and $\Pbotset$ are easy to evaluate, they can be directly translated into a stopping criterion. In other words, the algorithm stops when 
\begin{equation}\label{exploremstop}
 \Pbotset  \Ptopset > 1-\delta,
\end{equation}
is satisfied for the first time. 
Hence, the decision-making process is designed to optimize the stopping time. The final step consists in choosing the $\meanb$ value. We seek an optimal separator, i.e., the $\meanb$ value that enables fulfilling \cref{exploremstop} in minimal time. Introducing the total time needed to be spent on sampling each subset $\ttop$ and $\tbot$ to verify \cref{exploremstop} (regarding the current arms' mean state), $\meanb$ can be expressed as
\begin{equation}\label{explorem_meanbexp}
\meanb = \underset{\theta \in [\botmax, \topmin]}{\argmin} (\tbot(\meanb)  + \ttop(\meanb) |  \; \cref{exploremstop} \; \;  \mathrm{is\, verified}  ) .
\end{equation}

Unfortunately, there is no exact and tractable solution for \ref{explorem_meanbexp}. Hence, we seek a tractable approximation of ($\meanb$ , $\tbot(\meanb)$, $\ttop(\meanb)$). 
To this end, we isolated from the posterior distributions of the maximum and the minimum, respectively, of $\topsetmean$ and $\botsetmean$, the following closed-form approximations (Supplementary Material \cref{Supp:SecExplore}): 
\begin{equation}\label{explorem_meantop}
 \meanb  =  \frac{\epsilon}{2} + \topmin - \sqrt{\frac{\tk}{  \ttop(\meanb) }} \left[ \frac{2\ln(\tk) - \ln( - \ln(1 - \frac{\delta}{2}))}{ \sqrt{2\ln(\tk)}} \right],
\end{equation}
and 
\begin{equation}\label{explorem_meanbot}
\begin{split}
 \meanb  =  -&\frac{\epsilon}{2} + \botmax\\ &+   \sqrt{ \frac{\totk -\tk }{  \tbot(\meanb) }} \left[ \frac{2\ln(\totk -\tk) -  \ln( - \ln(1 - \frac{\delta}{2}))}{ \sqrt{2\ln(\totk -\tk)}} \right].
 \end{split}
\end{equation}

By inserting \cref{explorem_meanbot} and \cref{explorem_meantop} into \cref{explorem_meanbexp}, we obtain a closed system from which we derive a tractable expression of $\meanb$ given in \cref{sup:explorem_lastexprmeanb}. 
We refer to the Supplementary Material \cref{Supp:SecExplore} for a complete and detailed review of all the approximations and derivations. Briefly, using an extreme value analysis, we approximated $\Pbotset$ and $\Ptopset$ by Gumbel laws parametrized by $\topmin$, $\botmax$, $\tbot(\meanb)$ and $\ttop(\meanb)$. Next, they are evaluated when \cref{exploremstop} is verified, leading to \cref{explorem_meantop} and \cref{explorem_meanbot}.




Pseudocode for the resulting algorithm  referred to as \algonameexp{} is presented in \cref{alg:exp_algorithm}. 

\begin{algorithm}[htbp]
\DontPrintSemicolon
\caption{Algorithm for Explore-$m$ Gaussian bandits }
\label{alg:exp_algorithm}
\KwData{ $\totk$, $\tk$, $\epsilon$ , $\delta$,  $\hori>\totk$ (optional)}
Draw all arms once, observe rewards and update statistics for all arms, $t \gets \totk $\; 
$continue \gets \textbf{True}$\;
\While{$continue$}{
    \tcp*[f]{Stopping criteria}\;
    $\topset = \underset{I \subseteq \{1, \dots, n\}, \, |I| = m}{\argmax} \left( \meanh{i} \right)$ and  $\topmin = \min \topsetmeanh$\; 
    $\botset =  \{ \totk \} \setminus \topset $ and $\botmax = \max \botsetmeanh$\;
    Evaluate $\meanb$ using \cref{explorem_meanbexp}\;
    Evaluate $\Ptopset$ and $\Pbotset$ using \cref{explorePbot,explorePtop}\;
    \lIf{\cref{exploremstop} or $t=\hori$}{
         $continue \gets \textbf{False}$\;
    }
    \lElse{
    Pull $\armtp$ and observe $\rewardtp(\armtp)$\;
    Select $ \armt $ with \cref{explorem_armselect} \tcp*[f]{Arm selection}
    $\meanh{\armtp} \leftarrow \frac{\meanh{\armt}\loct{\armt} +\rewardtp(\armtp)}{ \loct{\armt} + 1} $\;
    $\loct{\armtp} \leftarrow \loct{\armt} + 1 $\tcp*[f]{Update statistics}
    }
}
\Return $\topset$\tcp*[f]{$\epsilon$-optimal set}
\end{algorithm}


\subsection{Results}
\begin{figure}[htpp]
\centering
\includegraphics[scale=0.7]{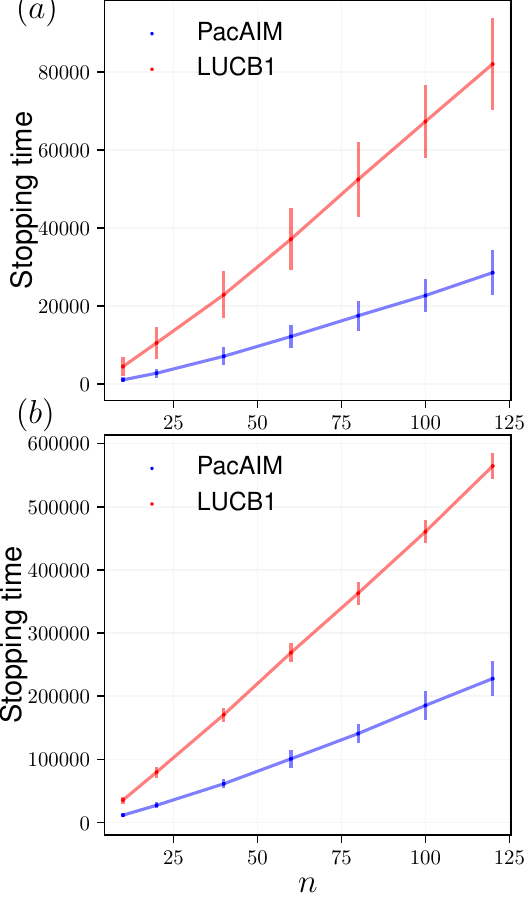}
\caption{Stopping time of Explore-$m$ algorithms for Gaussian rewards. The stopping time, i.e., the time when the stopping criteria for each algorithm is met, is measured for different numbers of arms $\totk$. In blue our algorithm, and in red LUCB1 given in \cite{kalyanakrishnanPACSubsetSelection2012a}.
The $\epsilon$-optimal subset size verifies $\tk= \totk/5$. In a) the mean rewards value are sampled from two disjoint ensemble $[0,0.8]$ and $[0.9,1]$ for the $\epsilon$-optimal subset, with confidence parameters $\epsilon=0.05$, $\delta=0.1$. b) Suboptimal mean rewards value are all set to $0.7$ while $\epsilon$-optimal to $0.8$, with confidence parameters  $\epsilon=0.05$, $\delta=0.02$. The time is averaged over $100$ runs with standard deviation indicated. See \cref{supp:Secnumericalexpe,supp:Secotheralgo} for numerical details and success probability rate.}
\label{fig:1}
\end{figure}

To evaluate the performance of \algonamesexp, we compare in \cref{fig:1} its stopping performance with the LUCB1 algorithm \cite{kalyanakrishnanPACSubsetSelection2012a}, a well-established algorithm in the PAC setting. The reward distributions are Gaussian with unknown mean values in $[0,1]$ and unit variance.

In the first experiment, the $\epsilon$-optimal subset is formed by mean rewards drawn uniformly from $[0.9,1]$, while the remaining mean rewards are drawn from $[0.0,0.8]$. To assess the algorithm's performance on more complex tasks, we also evaluate a scenario where all $\epsilon$-optimal rewards are set to $0.8$ and the remaining mean rewards are set to $0.7$. 
The stopping time is measured for different numbers of arms while maintaining the ratio $\tk = \totk/5$.

In both cases, \algonamesexp\ outperforms LUCB1, demonstrating the ability of our approximation to accurately capture the significant contributions of all arms near the $\meanb$ value. Notably, we emphasize that using a decision procedure directly tied to the stopping criteria enables the algorithm to sample all relevant arms efficiently, minimizing the stopping time. See the Supplemental Material \cref{supp:Secnumericalexpe,supp:Secotheralgo} for numerical details and additional experiments.

To conclude on the Explore-$m$ setting, we take a moment to highlight some additional observations. First, $\meanb$ depends on the current state of the game and evolves over time to adapt to the reward distributions. Next, the algorithm adjusts by sampling the arms that provide the most information relative to this updated separator value.

It is worth noting that our expression is designed for a subset consisting of multiple arms. For single-arm identification, one may replace \cref{explorem_meantop} by directly expressing $\meanb$ in terms of $\ttop$ using $\Prob \left( \topmin \geq \meanb - \frac{\epsilon}{2} | \Ht \right) = 1-\delta/2$. Furthermore, the expressions in \cref{explorem_meantop,explorem_meanbot} are derived for Gaussian rewards, but equivalent formulations can be obtained for various distribution families \cite{majumdarExtremeValueStatistics2020a}.

As a result, our algorithm is readily adaptable to more complex reward structures, including those with power-law tails, which lead to extreme value statistics scaling functions, such as Fréchet or Weibull distributions \cite{majumdarExtremeValueStatistics2020a}.
Altogether, these first results highlight the potential for extending information-based strategies to tackle more general pure exploration problems \citep{bubeckPureExplorationFinitelyarmed2011a, carpentierSimpleRegretInfinitely2015,rejwanTop$k$CombinatorialBandits2020}. The design of approximations and the choice of observables should be addressed in a problem-specific manner. However, our work can be easily adapted to threshold identification or gap identification problems  \citep{katariyaMaxGapBanditAdaptive2019, locatelliOptimalAlgorithmThresholding2016}, where the focus is on determining the gap between subsets. Additionally, our approach may be extended to identify all arms above a certain threshold or to target a specific fraction of arms, such as identifying all $\epsilon$-good arms or detecting outliers \citep{masonFindingAllTextbackslash2020, renExploring$k$Out2019,zhuangIdentifyingOutlierArms2017}. In all these examples, we expect information-based strategies to prove efficient in capturing the structure and correlations within the game \citep{bubeckPureExplorationFinitelyarmed2011a, carpentierSimpleRegretInfinitely2015}.

\section{Linear bandits}

\subsection{Setting and core insights}
Contextual bandits with linear payoffs is a fundamental paradigm in multi-armed bandit systems, underpinning key applications such as internet advertisement selection and personalized recommendations \cite{abeReinforcementLearningImmediate2003,chuContextualBanditsLinear2011,auerUsingConfidenceBounds2002}. This problem extends the classic bandit framework by introducing correlations between arms and allowing the set of arms to change over time \cite{lattimoreBanditAlgorithms2020,auerUsingConfidenceBounds2002, agrawalThompsonSamplingContextual2013, slivkinsContextualBanditsSimilarity2011} (see \cref{fig:schema} \textbf{c)}). For the stochastic linear problem, at each round $t$, the agent faces the decision set $\decset{} \in \Rd$ which may vary over time. From this set, it selects an action $\vecarm$ and observes a corresponding reward,
\begin{equation}
\rewardt = \langle \Astar, \vecarm \rangle + \xi_{t} ,
\end{equation}
where $\Astar$ defines the optimal direction and $\xi_{t}$ is modeled as Gaussian white noise in this study. Since the decision set may evolve over time, the losses are evaluated based on the best option available at any given time, leading to the random pseudo-regret:

\begin{equation}
 \hregret = \sum_{t=0}^\horin \underset{a \in \decset}{\max}  \langle \Astar, a - \vecarm \rangle 
\end{equation}
Hence, $\decset{}$ composition plays a crucial role in the setting. If $\decset{}$ forms a constant orthogonal basis, it simplifies to the classical bandit settings. The decision set may be finite or infinite, generated by a stochastic process, or part of an adversarial game adapting to previous choices, with $\decset{}$ potentially evolving rather than being drawn from a predefined set of decision sets \cite{lattimoreBanditAlgorithms2020}. This inherent diversity makes it difficult for linear bandit algorithms to be optimal across all settings \cite{lattimoreEndOptimismAsymptotic2017, combesMinimalExplorationStructured2017a}.
As in most bandit problems, the UCB algorithm emerges as a useful approach. It relies on a tuned confidence index associated with each arm to decide which arm to play. In the linear bandit case, the core algorithm is referred to as LinUCB, which achieves a logarithmic scaling of the regret when actions sufficiently close to $\Astar$ are available \cite{lattimoreBanditAlgorithms2020, daniStochasticLinearOptimization2008a, abbasi-yadkoriImprovedAlgorithmsLinear2011, chuContextualBanditsLinear2011, haoAdaptiveExplorationLinear2020, liContextualbanditApproachPersonalized2010, rusmevichientongLinearlyParameterizedBandits2010}. 
Otherwise, in the general case, the optimal regret is bounded by $O(d\ \sqrt{\hori})$. Beyond the UCB class, Thompson sampling, which relies on sampling mean rewards from a posterior distribution and choosing the arm with probability proportional to this posterior, 
has also been shown to achieve near-optimal regret bounds for stochastic contextual bandits with linear payoff functions \footnote{These results rely on an adjustment of the variance of the posterior distribution.} \cite{agrawalThompsonSamplingContextual2013,abeilleLinearThompsonSampling2017, kaufmannThompsonSamplingAsymptotically2012}. 
Optimal allocation matching strategies, on the other hand, rely on an estimate of $\Astar$ and  solve an optimization problem based on this estimate or perform exploration  to determine the next action
\citep{lattimoreEndOptimismAsymptotic2017, combesMinimalExplorationStructured2017a, haoAdaptiveExplorationLinear2020,  tirinzoniAsymptoticallyOptimalPrimaldual2020}. This class of strategies has been shown to achieve optimality when faced with a fixed action set \citep{haoAdaptiveExplorationLinear2020, combesMinimalExplorationStructured2017a}. Lastly, the information-directed sampling (IDS) principle leverages a measure of information gain from the optimal action, explicitly balanced by expected losses from exploration. By consistently using an information measure that captures the specific problem structure, IDS has been shown to achieve optimal regret scaling for linear bandits and other structured settings \citep{kirschnerAsymptoticallyOptimalInformationDirected2021, russoLearningOptimizeInformationDirected2014,  kirschnerInformationDirectedSampling2018, kirschnerInformationDirectedSampling2020}.
As first found in \cite{lattimoreEndOptimismAsymptotic2017}, a key insight is that a consistent strategy must allocate exploration time so that the gaps between all suboptimal actions and $\Astar$ are identified with high confidence. Otherwise, a purely greedy strategy solely based on $\Astar$ confidence estimation will fall short. This highlights the complexity arising from the variation of $\decset{}$.

We developed a new principle that relies almost exclusively on information acquisition to drive its decision policy. Our approach is rooted in the approximate information maximization (AIM) strategy, shown to be optimal in classic bandit settings \cite{barbier-chebbahApproximateInformationEfficient2023, barbier-chebbahApproximateInformationMaximization2023}.  The AIM principle involves selecting the arm that maximizes the information gained from a judiciously chosen observable of the game. This gain is measured through the associated entropy, with its increments approximated to identify the arm that will result in the largest entropy reduction. In \cite{barbier-chebbahApproximateInformationEfficient2023,barbier-chebbahApproximateInformationMaximization2023}, the entropy of the posterior distribution of the maximum mean reward value results in an optimal exploration-exploitation trade-off. However, it cannot be directly mapped to the linear case. By symmetry $\Amaxa$ and $-\Amaxa$ carry the same information. Thus, suboptimal actions can provide extensive information on $\Astar$ value but incur higher losses. Hence, solely focusing on acquiring more information on $\Astar$ results in over-exploration. We now present an adapted version of AIM to address this challenge. 



\subsection{Algorithm design}

Similar to Explore-$m$ bandits, we first define relevant notation. Let $ \bconfmt = \lambda I_d +  \sum_\horin^{t} \Ain \Ain^{\tr}$ be the covariance matrix of the posterior distribution of $\Astar$ according to a symmetric Gaussian prior parametrized by $\lambda$. 
We denote by $\thetaemp = \bconfmt^{-1}\sum_{\horin}^{t-1} \rewardn  \Ain$ its associated posterior mean (see Supplementary Material \cref{Supp:SecLinear} for detail). Here, $\tr$ and $^{-1}$ represent the transpose and inverse operators, respectively. The best available empirical arm in $\decset$ is $\Amaxa  = \underset{a \in \decset}{\max}  \langle \thetaemp, a \rangle $. 
We finally define a norm through $\|x\|_G = x' G x$ for $x \in \mathbb{R}^d$ and with $G \in \mathbb{R}^{d \times d}$ positive definite.


To adapt AIM, we opt to adjust the information gain depending on the selected arm. Arms will be selected to maximize the information regarding the value of  $\Astar$, weighted by the likelihood of the arm being suboptimal. 
In other words, the arms are evaluated not only by the information they provide but also regarding their relative gap with $\Amaxa$, quantified by the  likelihood that the best empirical choice $\Amaxa$ is better than $\Ait$, i.e., the probability that $\langle \Astar, \Amaxa - \Ait  \rangle > 0$. 
Finally, $\Amaxa$ is assigned to maximize the information regarding the $\Astar$ value. 
This procedure offers two main benefits: 1) it prevents over-exploration by adjusting the information gains from suboptimal arms. 2) It also involves a measure of the gaps between mean rewards, addressing the criteria set out by \cite{lattimoreEndOptimismAsymptotic2017}, which are also implicitly included in the information regarding $\Astar$. 

We now provide the analytical expression regarding the entropy increments relative to each choice (see Supplementary Material  \cref{Supp:SecLinear} for  details).

%
We first consider the information gain tied to $\Amaxa$. The posterior distribution of $\Astar$ is given by
\begin{equation}
p(\Astar = \theta | \Ht) = \frac{ \exp \left( - \frac{1}{2}(\theta- \thetaemp )^{\tr} \bconfmt (\theta- \thetaemp)  \right)} {\sqrt{(2 \pi)^d \det \bconfmt^{-1} }} ,
\end{equation}
where $\Ht$ represents the history of chosen arms and observed rewards. 
The associated entropy reads 
\begin{equation}\label{Sastar}
    S = \frac{1}{2} \ln \left(\det \bconfmt^{-1} \right)  + \frac{d}{2} \left[ 1 + \ln(2\pi) \right].
\end{equation}
The corresponding entropy increment, which quantifies the gain of information about $\Astar$, is then
\begin{equation}\label{DeltaSastar}
\Delta  S_{\Amaxa}  = - \frac{1}{2} \ln \left( 1 + \Amaxa^{\tr} \bconfmt^{-1} \Amaxa \right).
\end{equation}

We now consider the remaining arms, focusing on a given suboptimal arm $\Ait$. The  probability of observing  $\langle \Astar, \Amaxa - \Ait \rangle > 0$, termed $\probamai$, is given by 
%
%
\begin{equation}
 \probamai =  \frac{1}{2} \left[  1 + \erf \left( \frac{\Astar^{\tr} [\Amaxa - \Ait]}{ \sqrt{2 [\Amaxa - \Ait]^{\tr}  \bconfmt^{-1}  [\Amaxa - \Ait]}}\right)\right] ,
\end{equation}
with associated entropy
\begin{equation}\label{Sait}
S_{C,\Ait} = - \probamai \ln \probamai - (1- \probamai)  \ln(1-\probamai) .
\end{equation}

To achieve a simpler expression, we focus on the asymptotic form of \cref{Sait} when $\Amaxa$ is assumed to be the most probable optimal option. 
We denote this asymptotic form $w_{\Ait}$ such that $\quad S_{C,\Ait} \sim w_{\Ait} \quad \text{as} \quad \frac{\delta}{\sqrt{\Sigerf}} \rightarrow \infty$, which takes the form
\begin{equation}\label{DeltaSait}
\begin{split}
w_{\Ait} = \frac{\Meanerf}{\Sigerf^{\frac{3}{2}}} \frac{e^{-\frac{\Meanerf^2}{2 \Sigerf}}}{2 \sqrt{\pi}}, 
\end{split}
\end{equation}
%
 %
%
with $\Meanerf= \Astar^{\tr} [\Amaxa - \Ait] $ , $ \Sigerf = [\Amaxa - \Ait]^{\tr} \bconfmt^{-1}  [\Amaxa - \Ait]$ (see Supplementary Material \cref{Supp:SecLinear}). 
The weighted information gain relative to $\Ait$ thus reads
\begin{equation}\label{main:eqdeltadiltde}
\begin{split}
\Delta \tilde{S}_{\Ait} &= - w_{\Ait} \Delta  S_{\Ait}   \\
&=  - \frac{1}{2}  \frac{\Meanerf   }{\Sigerf^{\frac{3}{2}}}  \frac{e^{-\frac{\Meanerf^2}{2 \Sigerf}}}{2 \sqrt{\pi} }  \ln \left( 1 + \Ait^{\tr} \bconfmt^{-1} \Ait \right) .
\end{split}
\end{equation}


The pseudo-code for the resulting algorithm is presented in \cref{alg:linear_algorithm} below. 

\begin{algorithm}[htbp]
\DontPrintSemicolon
\caption{Algorithm for d-dimensional linear Gaussian bandits }
\label{alg:linear_algorithm}
Draw a first arm $ \Aio$, observe reward $\rewardo$, and update statistics: $  \bconfmu = \lambda I_d +   \Aio  \Aio^{\tr}$ and $\thetaempu = \bconfmu^{-1} \rewardo  \Aio$\;

\For(\tcp*[f]{Arm selection}){$t=1$ \KwTo $T$}{
   
    $ \Amaxa \gets  \underset{a \in \decset}{\max}  \langle \thetaemp, a \rangle $

    Evaluate $\Delta  S_{\Amaxa}$   following  \cref{DeltaSastar}
    
    \For{$\Ain \in \decset \backslash \{ \Amaxa \} $}{
        Evaluate $\Delta \tilde{S}_{\Ain}$ following \cref{main:eqdeltadiltde}\;
    }
    
   Select $ \Ait$ as 
   $\argmin \left( \Ain \in \decset \backslash \{  \Amaxa \} , \Amaxa : \Delta \tilde{S}_{\Ain}, \Delta S_{\Amaxa} \right ) $

   Pull $\Ait$ and observe  $\rewardt(\Ait)$
   
$ \bconfmtpo \gets \bconfmt +  \Ait \Ait^{\tr}$ and \\ $\thetaempo \gets \bconfmtpo^{-1} \sum_{\horin}^{t} \rewardn  \Ain$ \tcp*[f]{Update statistics}
}
\end{algorithm}


\subsection{Results}

Suboptimal arms are also evaluated regarding their gap with the best empirical arm available. As a result, the information gain expressed by \cref{DeltaSait} decreases exponentially with respect to this gap, making \cref{DeltaSastar} dominate in most cases, thereby preventing our algorithm from over-exploring.  In particular, it makes \eqref{DeltaSait} asymmetric with respect to $\Ait$, thereby enabling arms of opposite signs to be told apart. This fulfills the key challenge of preventing suboptimal arms from being sampled regarding the information gain they provide to close off optimal solutions. It also introduces the information relative to the gaps between current arms' means. Nonetheless, suboptimal arms are still evaluated based on the information they provide on the full system and under-probed directions, which stays the core of AIM. A preliminary theoretical guarantee can be assessed by considering the classic bandit setting, where $\decset$ is constant and forms an orthogonal basis. In this setting, \cref{DeltaSait,DeltaSastar} demonstrates asymptotic behavior similar to previous information maximization algorithms \cite{barbier-chebbahApproximateInformationEfficient2023} that have recently been proven optimal.
A second illustrative example involves an adversarial scenario where $\decset$ will suddenly start to display solely actions that have not been explored previously—untested directions, i.e. with high values of $\Amaxa^{\tr} \bconfmt^{-1} \Amaxa$. 
In such cases, the entropy increment \cref{DeltaSastar} should dominate, still accounting for exploration despite facing solely highly uncertain options.

\begin{figure}[htpp]
\centering
\includegraphics[scale=0.8]{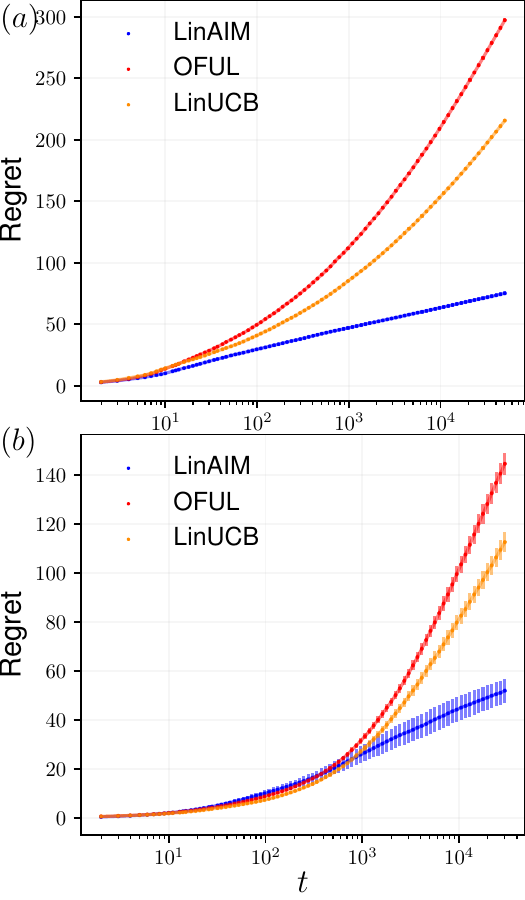}
\caption{ Mean regret for two linear bandit settings with Gaussian rewards. In blue our algorithm and in orange and red UCB algorithms adjusted to the linear settings. \textbf{(a)} Bandit setting with $10$ arms resampled at each time from a normal distribution in $\mathbb{R}^{10}$. \textbf{(b)} Toy problem borrowed from \cite{tirinzoniAsymptoticallyOptimalPrimaldual2020} with two contexts. Each context (detailed in the main text) is drawn with equal probability and $\xi=0.5$. Error bars show standard error around the mean. Details of the algorithms, simulations and a focus on standard deviations are provided in the Supplementary Material  \cref{supp:Secnumericalexpe,supp:Secotheralgo}.}
\label{fig:2}
\end{figure}

To evaluate the performance of \algonamelin , the average regret is analyzed in \cref{fig:2} for two complementary settings meant to cover a wide range of scenarios. In the first experiment, $\decset$ is formed by $10$ arms resampled at each time from a normal distribution in $\mathbb{R}^{10}$. This decision set exhibits temporal variability, with a sparse number of solutions available at each discrete time step. Here, \algonamelin{} performances match those of commonly used algorithms such as linUCB \cite{abbasi-yadkoriImprovedAlgorithmsLinear2011,lattimoreBanditAlgorithms2020}, slightly outperforming them. It showcases the algorithm's efficiency when confronted with a broad range of arms. To evaluate \algonamelin{}'s ability to leverage available information in suboptimal solutions, we consider a toy setting defined in \cite{tirinzoniAsymptoticallyOptimalPrimaldual2020} originally from \cite{lattimoreEndOptimismAsymptotic2017}. The agent faces two fixed sets $\decset$ formed by three arms in $\mathbb{R}^{3}$, each set having an equal probability of being presented to the agent. The first set is \(\{[1, 0, 0], [0, 1, 0], [1-\xi, 2\xi, 0]\}\), the second set \(\{[0, 0.6, 0.8], [0, 0, 1], [0, \xi/10, 1-\xi]\}\), and finally \(\Astar = [1, 0, 1]\) with $\xi$ small. The key insight is that optimistic strategies are rarely selected $[0, 1, 0]$  $[0, 0.6, 0.8]$ in the first and second sets due to the substantial gap. They thereby incur significant regret in learning which of the remaining arms is optimal. Conversely, an asymptotically optimal strategy strategically allocates more selections to the suboptimal arms, facilitating information acquisition to accurately identify $\Astar$, hence circumventing regret that scales with accurately $\xi$. Here also, our algo shows strong performance similar to state-of-the-art algorithms.
Such performances are quite convincing, and our algorithm demonstrates performance comparable to recent algorithms such as information-directed sampling \cite{kirschnerAsymptoticallyOptimalInformationDirected2021, kirschnerInformationDirectedSampling2020}, SOLID \cite{tirinzoniAsymptoticallyOptimalPrimaldual2020}, OSSB \cite{combesMinimalExplorationStructured2017a} or OAM \cite{haoAdaptiveExplorationLinear2020} . Although our algorithm shares similarities with information-directed sampling by adjusting the information gain, our design and analytic expressions are notably simpler than the methods cited above. It notably reduces the implementation costs and doesn't require working with finite arm sets. Nevertheless, linear bandit setting is highly complex, in particular in the adversarial regime where the decision set can be specifically designed to expose the algorithm's flaws. Hence, we must confirm the empirical performance of \algonamelin{} by providing theoretical guarantees, and we may identify settings where its effectiveness is limited. Nonetheless, its design is sufficiently simple to allow substantial extensions to meet these theoretical criteria. In particular, suboptimal actions could also be evaluated based on their ability to provide information about all available gaps, or specifically the gaps between $\thetaemp$ and all suboptimal arms. This initial design lays the foundation for a deeper exploration of information-based approaches in linear bandits.


\section{Many-armed bandits}
\subsection{Setting and core insight}

We now consider the many-armed setting as introduced in \cite{bayatiUnreasonableEffectivenessGreedy2020}. The major difference with the classic bandit problem is that the game's duration is finite and known in advance. It aims to cover problems in which the number of actions is too large for exhaustive search [\cref{fig:schema} \textbf{d)}]. It underpins key applications that require multiple testing, such as encountered in medical trials \cite{jamiesonBanditApproachMultiple2018, villarBanditsStrategiesEvaluated2018, villarMultiarmedBanditModels2015}. The agent must carefully define patient testing groups, balancing ethical standards and treatment costs while ensuring maximum information gain on trial efficacy. 
In this framework, in each round $t$ (where $t = 1,..,T$), the agent selects an arm $\armt \in \kset =\{1,..,\kv\}$, where $\kv$ represents the total number of arms. The chosen arm yields a stochastic reward $\rewardt(\armt)$, drawn from a distribution of unknown mean $\meant$. We assume a Bayesian framework where the mean reward for each arm is drawn from a common prior distribution $\muprior$ with a finite support $[\mubmin, \mubmax]$. Furthermore, we assume that the reward distributions for all arms belong to the same family, parameterized solely by their means.  For instance, the rewards might follow Gaussian distributions of known fixed variance $\sigma^2$ where the mean of each arm is drawn uniformly from the interval $[0,1]$. Given a time horizon $\hori$, the agent's goal is to minimize its expected pseudo-regret over $\hori$ rounds in a single experiment. The pseudo-regret is here given by
\begin{equation}
\regretT = \hori \, \maxK \, \mean{i} - \sum_{t=1}^\hori \E{\meant} .
\end{equation}
This measures the gap between the cumulative reward from always choosing the best arm and the agent's expected reward over $\hori$ rounds. The expectation reflects that the agent's policy, informed by past rewards, is affected by their randomness. Ultimately, the agent aims to minimize the Bayesian expected regret, 
\begin{equation}
\BR = \E{\regretT} .
\end{equation}
Here, the expectation is taken over the prior distribution of the mean rewards, representing the agent's performance averaged over all possible mean reward configurations. The key distinction from previous settings is that the agent operates with a finite budget and has prior knowledge of the total number of rounds $T$. We focus on scenarios where $\kv > \sqrt{\hori}$, yielding too little time for the agent to extensively sample all the arms or fully exploit the best ones. To achieve an efficient strategy, the agent must sample only a subset of size $m \sim O(\sqrt{\hori})$ arms, enabling the algorithm to find a promising solution without inducing excessive exploration \cite{bayatiUnreasonableEffectivenessGreedy2020}. Beyond this scale, further exploration becomes too costly compared to leveraging the best empirical arms. Conversely, before this scale, the sampled arms have a higher probability of inducing significant regret, making early exploitation less effective \cite{bayatiUnreasonableEffectivenessGreedy2020}. 

Multiple strategies attain the optimal scaling of $\BR \approx O(\sqrt{\hori})$ \footnote{The lower bounds holds for sufficiently high $\hori,\kv$ values and $1$-regular $\muprior$ ($\mubmax=1$), meaning that $\mathcal{P}_\muprior[ \mu > 1- \epsilon] = O(\epsilon)$ as $\epsilon$ goes to $0$; see \cite{bayatiUnreasonableEffectivenessGreedy2020} for theorem details.}. Notably, Subsampled Greedy (SS-Greedy) relies on an ad hoc selection of a subset of fixed size ($O(\sqrt{\hori})$) to limit exploration and has been shown to be optimal for Bernoulli rewards. While such strategies achieve optimal asymptotic scaling, they do not allow the algorithm to adapt to the rewards observed during its execution.  
Hence, we aim to use a physics-based approach that naturally recovers the $O(\sqrt{\hori})$ scaling through a functional expression, while maintaining flexibility and both short- and long-time efficiency.

\subsection{Algorithm design}
We propose an algorithm that leverages the minimization of a global functional based on underlying physics principles, using as a proxy variable the expected upcoming regret, i.e., the regret accumulated from time $t$ to $T$. 
Our policy works as follows: at each time step, it evaluates the expected upcoming regret for two distinct decision sequences and selects the one with the lower expected upcoming regret. The two sequences  are: 1) pull the current best empirical arm, denoted $\maxa$, until the end, or 2) try a new untested arm  until it eventually underperforms compared to $\maxa$, at which point $\maxa$ is chosen. 
Although these strategies recommend actions for the entire remaining time, it does not mean the algorithm will follow them in subsequent steps, as the choice between them is reassessed at each time step. Before evaluating both upcoming regrets, we need to introduce several definitions. When clear from context, we omit the dependence on $t$ for simplicity. The number of times arm $i$ has been pulled is denoted by $\loct{i}$, and its associated empirical mean by $\meanh{i}$. The number of different arms tested by time $t$ is labeled $\nv$, forming the set $\nset=\{k_1, k_2, \ldots, \nv\}$.  The best empirical arm at time $t$ is labeled $\maxa$, with its corresponding empirical mean $\mumaxt = \maxn , \meanh{i}(t)$. We also define the remaining time as $\retime = T-t$. Last, the expected maximum mean value is denoted $\mustar = \Es{\maxK\, \mean{i}}{\muprior} $.  

We now derive approximations for the regret in both cases, see the Supplementary Material \cref{Supp:SecMany} for further details and discussion of all approximations. For 1) (fully exploit $\maxa$ for the next $\tau$ turns) the approximated upcoming expected regret is equal to
\begin{equation}
\Rgr = \retime (\mustar - \mumaxt)
\end{equation}
Here, we approximated the expected value of the current best empirical arm by $\mumaxt$. 
For 2) (testing a new arm) the upcoming expected regret is approximated as 
\begin{equation}\label{eqgeneralmany}
\begin{split}
\Rexp &=  \int_{\mumaxt}^{\mubmax}  \muprior(\mu) (\mustar - \mu) \tau\df \mu  \: \: + \\ &\int_{\mubmin}^{\mumaxt} \muprior(\mu) \left[  (\mustar-  \mu)\tfpt  + (\mustar - \mumaxt) \left( \tau - \tfpt \right)  \right] \df \mu.
\end{split}
\end{equation}
The first integral captures cases where the true mean reward of the new arm is higher than $\mumaxt$. 
Roughly, to obtain this expression we neglect cases where the empirical mean of a promising arm  never drops below $\mumaxt$. 
The second integral corresponds to cases where the true mean reward value of the new arm is lower than $\mumaxt$. In this case, the new arm will be pulled for an average time $\tfpt$ until it becomes suboptimal, which happens when its empirical mean falls below $\mumaxt$. 
This first phase yields the first term, following which the algorithm reverts to exploiting the best empirical arm, as captured by the second term. 

Finally, the decision procedure evaluates  
\begin{equation}
\begin{split}
\Delta &= \Rexp -\Rgr \\
 &=  \retime \int_{\mumaxt}^{\mubmax} \muprior(\mu)(\mumaxt - \mu) \df \mu\\  & \hspace{3cm}+ \int_{\mubmin}^{\mumaxt} \muprior(\mu) (\mumaxt -\mu) \tfpt \df \mu ,
 \end{split}
\end{equation}
and follows the strategy associated with the lowest upcoming regret. 
To get a universal and simpler expression, we employ the scaling ansatz that the typical number of steps to assess suboptimality is inversely proportional to the mean differences, $\tfpt \propto (\mumaxt - \mu)^{-1}$ (refer to Supplementary Material\cref{Supp:SecMany} for additional details). 

For a uniform prior under $[0,1]$, we obtain the following  expression for $\Delta$
\begin{equation}\label{eq:manyDeltaf}
\Delta = c \mumaxt - \retime \frac{(1-\mumaxt)^2}{2} ,
\end{equation}
where $c$ is a free parameter, which we set to $1$. 

The pseudo-code for the resulting algorithm referred to as approximate expected regret minimization (\algonamesemany) is presented in \cref{alg:many_algorithm} below. 

\begin{algorithm}[htbp]
\DontPrintSemicolon
\caption{Algorithm for $k$ Gaussian arms with finite horizon $\hori$}
\label{alg:many_algorithm}
Draw a first arm $A_0$, observe reward $\rewardo$, and update statistics: $\meanh{0} = \rewardo$, $\nz = 1$\;

\For(\tcp*[f]{Arm selection}){$t=1$ \KwTo $\hori$}{
   
    $ \maxa \gets \underset{ i \in \nset}{\mathrm{argmax}} \, \meanh{i}(t)$\;
    
    \lIf{$\nv = k$}{
        $\armt \gets \maxa$
    }
    \Else{
        Evaluate $\Delta$ following \cref{eq:manyDeltaf}\;
        \lIf{$\Delta \leq 0$}{
            $\armt \gets \maxa$
        }
        \lElse{
            $\armt \gets \nv + 1$, $\nvp \gets \nv + 1$
        }
        Pull $\armt$ and observe $\rewardt(\armt)$\;
    }

$\meanh{\armtp} \leftarrow \frac{\meanh{\armt}\loct{\armt} +\rewardt(\armt)}{ \loct{\armt} + 1} $,\tcp*[f]{Update statistics}
$\loct{\armtp} \leftarrow \loct{\armt} + 1 $
}
\end{algorithm}

Lets us briefly comment on our algorithm design. When $\Delta$ is positive, the current best empirical arm is chosen, as it involves a lower expected cost. Conversely, if $\Delta$ is negative, a new untested arm is pulled. Note that the algorithm behaves greedily when no more arms remain to be tested, although such cases should rarely be encountered in many-armed settings. These simplifications aim to derive a straightforward and tractable expression, while still leveraging a well-defined global quantity to minimize. Here, since the time horizon is too short for leveraging a full exploration and identification of the best arm, the need to acquire more information must be significantly limited in the decision process. Therefore, here the relevant information lies in the time required to assess that a new arm is sub-optimal and is quantified by $\tfpt$.


\begin{figure}[htpp]
\centering
\includegraphics[scale=0.8]{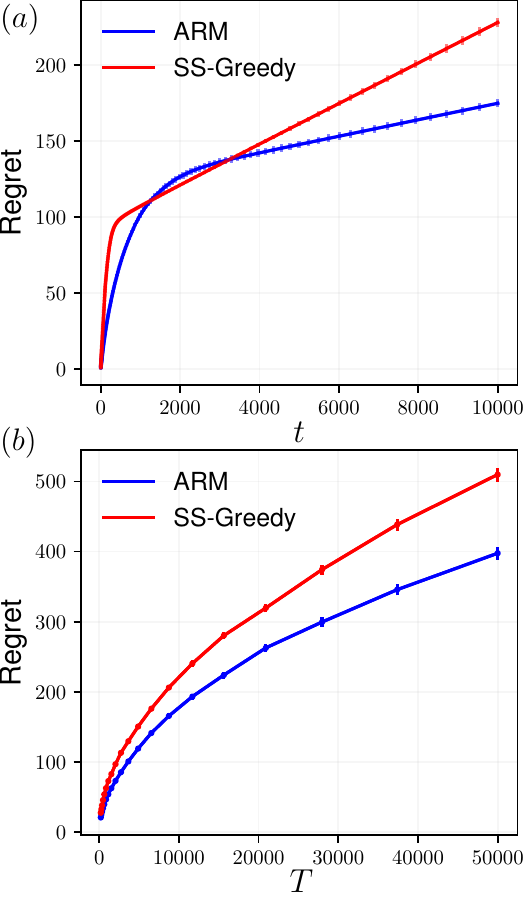}
\caption{Mean regret for the many-armed bandit problem with Bernoulli rewards and a uniform prior on $[0,1]$. \textbf{(a)} The regret growth of our algorithm (\algonamesemany), in blue, is observed until horizon $\hori$, i.e the stopping time, is reached $\hori=10000$ ($K=5000 \gg \sqrt{\hori}$ and $c=1$). The regret is averaged over $2000$ runs and standard error are indicated  (see Supplementary Material \cref{supp:Secnumericalexpe,supp:Secotheralgo} for details on the numerical settings). 
Its performance is compared to the ss-greedy algorithm, in red. The final slopes of both algorithms differs indicating that the selected arm for most of the exploitation is less efficient than the one found for our algorithm. Additionally, our algorithm presents a slower initial slope because exploration and exploitation are already mixed. \textbf{(b)} Mean regret performance when at the stopping time (when the horizon is reached),  for distinct games with varying horizon length ($K=\hori \gg \sqrt{\hori}$ and $c=1$). The regret is averaged over $2000$ runs with error bars indicating standard error on the mean. Details of the algorithms and simulations, as well as a focus on standard deviations, are provided in the Supplementary Material  \cref{supp:Secnumericalexpe,supp:Secotheralgo}.}
\label{fig:3pres}
\end{figure}

\subsection{Results}

To assess the efficiency of \algonamesemany{} (\cref{alg:many_algorithm}), its average regret is compared with the state-of-the-art SS-epsilon greedy algorithm (see \cite{bayatiUnreasonableEffectivenessGreedy2020} and \cref{supp:Secotheralgo} in the Supplementary Material for implementation details).  Notably, SS-epsilon greedy has been shown to achieve an optimal regret scaling of $O(\sqrt{\hori})$ for $\kv \gg \sqrt{\hori}$. 
Figure \ref{fig:3pres} (a) shows the regret in the many-armed bandit setting with a finite horizon and Bernoulli rewards, where our algorithm outperforms the SS-greedy algorithm. This is also confirmed for several horizon durations [\cref{fig:3pres}(b)].
This numerical result outlines that the optimal balance between exploration and exploitation is already embedded in the underlying physical structure of our algorithm. In contrast, SS-greedy manually imposes a first exploration phase of constant duration (namely $\sqrt{\hori}$) to ensure optimal scaling, followed by a greedy strategy, leading to disjoint exploration-exploitation phases. 
This is exemplified by \algonamesemany{} displaying a smoother initial regret curve compared to SS-greedy.  \algonamesemany{} alternates between exploration and exploitation until a sufficiently promising arm is found, adapting its exploration amount around the fixed $\sqrt{\hori}$ ratio if a sufficiently promising arm is found earlier or later~\footnote{Note that a sequential strategy has been proposed in \cite{bayatiUnreasonableEffectivenessGreedy2020}, but this strategy still depends on a fixed bound on the subset size.}. 
Notably, in \cref{fig:3pres} (a), the selected arm is found later by \algonamesemany{} than by the SS-greedy policy but outperforms the one found by SS-greedy on average, as indicated by the distinct final slopes of the regret closer to the horizon. 

The empirically observed optimality of \algonamesemany{} is supported by a heuristic scaling argument for when the transition between exploration and exploitation likely occurs (see Supplementary Material \cref{Supp:SecMany} for additional elements).  
Briefly, the expected value of $\Delta$ [\cref{eq:manyDeltaf}] crosses over from positive to negative values for $\retime(1-\mumaxt) \sim O(1)$. 
Additionally, the typical value of $\mumaxt$ scales with $t$ as $(1-\mumaxt) \sim 1/t$ since it is inversely proportional to the number of tested arms, which is in turn proportional to $t$ in the exploration phase (before the crossover from positive to negative $\Delta$). 
Combining these scales implies that $t \sim O(\sqrt{\hori})$, thereby explaining the optimal scaling of \algonamesemany{}'s regret. 
Note, however, that this argument does not constitute a rigorous proof, which would need to account for the influence of rare events, such as failing to identify an arm for exploration within $O(\sqrt{\hori})$ time. 

We conclude this section by several remarks.  
First, \algonamesemany{} relies on its ability to combine its knowledge of the remaining time and the available information on $\Gamma$ to optimize its decision process. Notably, \cref{eqgeneralmany} applies to more general prior distributions. If the prior is not provided to the agent, \algonamesemany{} could be adapted to add an empirical evaluation of $\Gamma$, in particular close to its upper bound. Similar to SS-greedy,  \algonamesemany{} exploits by selecting the arm with the best empirical mean, which still allows identifying a sufficiently efficient arm within the subset of tested arms. To enhance the exploitation component and leverage prior knowledge, one may consider selecting the arm with the highest probability of being optimal, rather than the highest posterior mean. This slight difference is crucial, as the algorithm might otherwise fall into exploiting arms with better averages but a higher probability of being suboptimal compared to other explored arms. To implement such a strategy, one could replace $\mumaxt$ in \cref{eq:manyDeltaf} by the expected mean reached by playing this alternative strategy given the current tested arms. But such an alternative remains more complex than using the empirical mean as we showed already provides efficient results. This approach could be considered for future developments. This flexibility should favor the extension and adaptability of \algonamesemany{} to more complex priors and bandit problems.


\section{Conclusion}

This study presents a generalisation of the information maximization principle and physics-based representations to a varied selection of structured bandit games. Our study addresses three distinct problems within the multi-armed bandit domain: many-armed bandits, linear bandits, and pure exploration. Each of these settings represents a unique facet of the broad field of bandit games, and each requires a tailored approach to account for their underlying structures and achieve the right balance between exploration and exploitation. 
To this end, for each setting, we focused on a specific observable from which we extracted a simplified and analytical functional at the core of the decision scheme. 
Numerical experiments in the paradigmatic versions of these games showed the robustness and efficiency of our method. Further research should focus on deriving rigorous theoretical bounds tailored to each bandit setting. Our method goes beyond classical confidence intervals for individual arms by leveraging a functional representation to guide decision-making. This core architecture is designed to make our method more adaptable to the diverse variants that may be encountered in real-world scenarios. Each algorithm examined above could be adapted to address more intricate variations of the corresponding problems, especially when dealing with reward or prior distributions exhibiting power-law tails, which induce extreme values. Future work will focus on searching for demonstrations of optimality (such as the ones derived for AIM) and systematic approaches to devise information-based approaches to decision-making processes.

\bibliographystyle{vincent} 
\bibliography{StructuredBandit.bib}

\begin{thebibliography}{101}
\expandafter\ifx\csname natexlab\endcsname\relax\def\natexlab#1{#1}\fi
\expandafter\ifx\csname bibnamefont\endcsname\relax
  \def\bibnamefont#1{#1}\fi
\expandafter\ifx\csname bibfnamefont\endcsname\relax
  \def\bibfnamefont#1{#1}\fi
\expandafter\ifx\csname citenamefont\endcsname\relax
  \def\citenamefont#1{#1}\fi
\expandafter\ifx\csname url\endcsname\relax
  \def\url#1{\texttt{#1}}\fi
\expandafter\ifx\csname urlprefix\endcsname\relax\def\urlprefix{URL }\fi
\providecommand{\bibinfo}[2]{#2}
\providecommand{\eprint}[2][]{\url{#2}}

\bibitem[{\citenamefont{Sutton and Barto}(1998)}]{suttonReinforcementLearningIntroduction1998a}
\bibinfo{author}{\bibfnamefont{R.~S.} \bibnamefont{Sutton}} \bibnamefont{and} \bibinfo{author}{\bibfnamefont{A.~G.} \bibnamefont{Barto}}, \emph{\bibinfo{title}{Reinforcement {{Learning}}: {{An Introduction}}}}, Adaptive {{Computation}} and {{Machine Learning}} Series (\bibinfo{publisher}{A Bradford Book}, \bibinfo{address}{Cambridge, MA, USA}, \bibinfo{year}{1998}), ISBN \bibinfo{isbn}{978-0-262-19398-6}.

\bibitem[{\citenamefont{Gold and Shadlen}(2007)}]{goldNeuralBasisDecision2007}
\bibinfo{author}{\bibfnamefont{J.~I.} \bibnamefont{Gold}} \bibnamefont{and} \bibinfo{author}{\bibfnamefont{M.~N.} \bibnamefont{Shadlen}}, \bibinfo{journal}{Annu. Rev. Neurosci.} \textbf{\bibinfo{volume}{30}}, \bibinfo{pages}{535} (\bibinfo{year}{2007}).

\bibitem[{\citenamefont{Jepma and Nieuwenhuis}(2011)}]{jepmaPupilDiameterPredicts2011}
\bibinfo{author}{\bibfnamefont{M.}~\bibnamefont{Jepma}} \bibnamefont{and} \bibinfo{author}{\bibfnamefont{S.}~\bibnamefont{Nieuwenhuis}}, \bibinfo{journal}{Journal of Cognitive Neuroscience} \textbf{\bibinfo{volume}{23}}, \bibinfo{pages}{1587} (\bibinfo{year}{2011}).

\bibitem[{\citenamefont{Kim et~al.}(2009)\citenamefont{Kim, Sul, Huh, Lee, and Jung}}]{kimRoleStriatumUpdating2009}
\bibinfo{author}{\bibfnamefont{H.}~\bibnamefont{Kim}}, \bibinfo{author}{\bibfnamefont{J.~H.} \bibnamefont{Sul}}, \bibinfo{author}{\bibfnamefont{N.}~\bibnamefont{Huh}}, \bibinfo{author}{\bibfnamefont{D.}~\bibnamefont{Lee}}, \bibnamefont{and} \bibinfo{author}{\bibfnamefont{M.~W.} \bibnamefont{Jung}}, \bibinfo{journal}{J Neurosci} \textbf{\bibinfo{volume}{29}}, \bibinfo{pages}{14701} (\bibinfo{year}{2009}).

\bibitem[{\citenamefont{Vergassola et~al.}(2007)\citenamefont{Vergassola, Villermaux, and Shraiman}}]{vergassolaInfotaxisStrategySearching2007}
\bibinfo{author}{\bibfnamefont{M.}~\bibnamefont{Vergassola}}, \bibinfo{author}{\bibfnamefont{E.}~\bibnamefont{Villermaux}}, \bibnamefont{and} \bibinfo{author}{\bibfnamefont{B.~I.} \bibnamefont{Shraiman}}, \bibinfo{journal}{Nature} \textbf{\bibinfo{volume}{445}}, \bibinfo{pages}{406} (\bibinfo{year}{2007}).

\bibitem[{\citenamefont{Murlis et~al.}(1992)\citenamefont{Murlis, Elkinton, and Card{\'e}}}]{murlisOdorPlumesHow1992}
\bibinfo{author}{\bibfnamefont{J.}~\bibnamefont{Murlis}}, \bibinfo{author}{\bibfnamefont{J.~S.} \bibnamefont{Elkinton}}, \bibnamefont{and} \bibinfo{author}{\bibfnamefont{R.~T.} \bibnamefont{Card{\'e}}}, \bibinfo{journal}{Annual Review of Entomology} \textbf{\bibinfo{volume}{37}}, \bibinfo{pages}{505} (\bibinfo{year}{1992}).

\bibitem[{\citenamefont{Masson et~al.}(2009)\citenamefont{Masson, Bechet, and Vergassola}}]{massonChasingInformationSearch2009b}
\bibinfo{author}{\bibfnamefont{J.-B.} \bibnamefont{Masson}}, \bibinfo{author}{\bibfnamefont{M.~B.} \bibnamefont{Bechet}}, \bibnamefont{and} \bibinfo{author}{\bibfnamefont{M.}~\bibnamefont{Vergassola}}, \bibinfo{journal}{J. Phys. A: Math. Theor.} \textbf{\bibinfo{volume}{42}}, \bibinfo{pages}{434009} (\bibinfo{year}{2009}).

\bibitem[{\citenamefont{Wang and Jegelka}(2017)}]{wangMaxvalueEntropySearch2017}
\bibinfo{author}{\bibfnamefont{Z.}~\bibnamefont{Wang}} \bibnamefont{and} \bibinfo{author}{\bibfnamefont{S.}~\bibnamefont{Jegelka}}, in \emph{\bibinfo{booktitle}{Proceedings of the 34th {{International Conference}} on {{Machine Learning}}}} (\bibinfo{publisher}{PMLR}, \bibinfo{year}{2017}), pp. \bibinfo{pages}{3627--3635}, ISSN \bibinfo{issn}{2640-3498}.

\bibitem[{\citenamefont{Russo and Van~Roy}(2014)}]{russoLearningOptimizeInformationDirected2014}
\bibinfo{author}{\bibfnamefont{D.}~\bibnamefont{Russo}} \bibnamefont{and} \bibinfo{author}{\bibfnamefont{B.}~\bibnamefont{Van~Roy}}, in \emph{\bibinfo{booktitle}{Advances in {{Neural Information Processing Systems}}}}, edited by \bibinfo{editor}{\bibfnamefont{Z.}~\bibnamefont{Ghahramani}}, \bibinfo{editor}{\bibfnamefont{M.}~\bibnamefont{Welling}}, \bibinfo{editor}{\bibfnamefont{C.}~\bibnamefont{Cortes}}, \bibinfo{editor}{\bibfnamefont{N.}~\bibnamefont{Lawrence}}, \bibnamefont{and} \bibinfo{editor}{\bibfnamefont{K.~Q.} \bibnamefont{Weinberger}} (\bibinfo{publisher}{Curran Associates, Inc.}, \bibinfo{year}{2014}), vol.~\bibinfo{volume}{27}.

\bibitem[{\citenamefont{Lindner et~al.}(2022)\citenamefont{Lindner, Turchetta, Tschiatschek, Ciosek, and Krause}}]{lindnerInformationDirectedReward2022}
\bibinfo{author}{\bibfnamefont{D.}~\bibnamefont{Lindner}}, \bibinfo{author}{\bibfnamefont{M.}~\bibnamefont{Turchetta}}, \bibinfo{author}{\bibfnamefont{S.}~\bibnamefont{Tschiatschek}}, \bibinfo{author}{\bibfnamefont{K.}~\bibnamefont{Ciosek}}, \bibnamefont{and} \bibinfo{author}{\bibfnamefont{A.}~\bibnamefont{Krause}}, \emph{\bibinfo{title}{Information {{Directed Reward Learning}} for {{Reinforcement Learning}}}} (\bibinfo{year}{2022}), \eprint{2102.12466}.

\bibitem[{\citenamefont{Houthooft et~al.}(2017)\citenamefont{Houthooft, Chen, Duan, Schulman, Turck, and Abbeel}}]{houthooftVIMEVariationalInformation2017}
\bibinfo{author}{\bibfnamefont{R.}~\bibnamefont{Houthooft}}, \bibinfo{author}{\bibfnamefont{X.}~\bibnamefont{Chen}}, \bibinfo{author}{\bibfnamefont{Y.}~\bibnamefont{Duan}}, \bibinfo{author}{\bibfnamefont{J.}~\bibnamefont{Schulman}}, \bibinfo{author}{\bibfnamefont{F.~D.} \bibnamefont{Turck}}, \bibnamefont{and} \bibinfo{author}{\bibfnamefont{P.}~\bibnamefont{Abbeel}}, \emph{\bibinfo{title}{{{VIME}}: {{Variational Information Maximizing Exploration}}}} (\bibinfo{year}{2017}), \eprint{1605.09674}.

\bibitem[{\citenamefont{Ding et~al.}(2019)\citenamefont{Ding, Li, and Liu}}]{dingInteractiveAnomalyDetection2019}
\bibinfo{author}{\bibfnamefont{K.}~\bibnamefont{Ding}}, \bibinfo{author}{\bibfnamefont{J.}~\bibnamefont{Li}}, \bibnamefont{and} \bibinfo{author}{\bibfnamefont{H.}~\bibnamefont{Liu}}, in \emph{\bibinfo{booktitle}{Proceedings of the {{Twelfth ACM International Conference}} on {{Web Search}} and {{Data Mining}}}} (\bibinfo{publisher}{Association for Computing Machinery}, \bibinfo{address}{New York, NY, USA}, \bibinfo{year}{2019}), {{WSDM}} '19, pp. \bibinfo{pages}{357--365}, ISBN \bibinfo{isbn}{978-1-4503-5940-5}.

\bibitem[{\citenamefont{Zhang et~al.}(2015)\citenamefont{Zhang, Martinez, and Masson}}]{zhangMultiRobotSearchingSparse2015c}
\bibinfo{author}{\bibfnamefont{S.}~\bibnamefont{Zhang}}, \bibinfo{author}{\bibfnamefont{D.}~\bibnamefont{Martinez}}, \bibnamefont{and} \bibinfo{author}{\bibfnamefont{J.-B.} \bibnamefont{Masson}}, \bibinfo{journal}{Frontiers in Robotics and AI} \textbf{\bibinfo{volume}{2}} (\bibinfo{year}{2015}).

\bibitem[{\citenamefont{{Hernandez-Lobato} et~al.}(2016)\citenamefont{{Hernandez-Lobato}, {Hernandez-Lobato}, Shah, and Adams}}]{hernandez-lobatoPredictiveEntropySearch2016}
\bibinfo{author}{\bibfnamefont{D.}~\bibnamefont{{Hernandez-Lobato}}}, \bibinfo{author}{\bibfnamefont{J.}~\bibnamefont{{Hernandez-Lobato}}}, \bibinfo{author}{\bibfnamefont{A.}~\bibnamefont{Shah}}, \bibnamefont{and} \bibinfo{author}{\bibfnamefont{R.}~\bibnamefont{Adams}}, in \emph{\bibinfo{booktitle}{Proceedings of {{The}} 33rd {{International Conference}} on {{Machine Learning}}}}, edited by \bibinfo{editor}{\bibfnamefont{M.~F.} \bibnamefont{Balcan}} \bibnamefont{and} \bibinfo{editor}{\bibfnamefont{K.~Q.} \bibnamefont{Weinberger}} (\bibinfo{publisher}{PMLR}, \bibinfo{address}{New York, New York, USA}, \bibinfo{year}{2016}), vol.~\bibinfo{volume}{48} of \emph{\bibinfo{series}{Proceedings of {{Machine Learning Research}}}}, pp. \bibinfo{pages}{1492--1501}.

\bibitem[{\citenamefont{Martinez et~al.}(2014)\citenamefont{Martinez, Arhidi, Demondion, Masson, and Lucas}}]{martinezUsingInsectElectroantennogram2014}
\bibinfo{author}{\bibfnamefont{D.}~\bibnamefont{Martinez}}, \bibinfo{author}{\bibfnamefont{L.}~\bibnamefont{Arhidi}}, \bibinfo{author}{\bibfnamefont{E.}~\bibnamefont{Demondion}}, \bibinfo{author}{\bibfnamefont{J.-B.} \bibnamefont{Masson}}, \bibnamefont{and} \bibinfo{author}{\bibfnamefont{P.}~\bibnamefont{Lucas}}, \bibinfo{journal}{JoVE (Journal of Visualized Experiments)} p. \bibinfo{pages}{e51704} (\bibinfo{year}{2014}).

\bibitem[{\citenamefont{Card{\'e}}(2021)}]{cardeNavigationWindbornePlumes2021}
\bibinfo{author}{\bibfnamefont{R.~T.} \bibnamefont{Card{\'e}}}, \bibinfo{journal}{Annual Review of Entomology} \textbf{\bibinfo{volume}{66}}, \bibinfo{pages}{317} (\bibinfo{year}{2021}).

\bibitem[{\citenamefont{Cohen et~al.}(2007)\citenamefont{Cohen, McClure, and Yu}}]{cohenShouldStayShould2007}
\bibinfo{author}{\bibfnamefont{J.~D.} \bibnamefont{Cohen}}, \bibinfo{author}{\bibfnamefont{S.~M.} \bibnamefont{McClure}}, \bibnamefont{and} \bibinfo{author}{\bibfnamefont{A.~J.} \bibnamefont{Yu}}, \bibinfo{journal}{Philosophical Transactions of the Royal Society B: Biological Sciences} \textbf{\bibinfo{volume}{362}}, \bibinfo{pages}{933} (\bibinfo{year}{2007}).

\bibitem[{\citenamefont{Doya}(2007)}]{doyaBayesianBrainProbabilistic2007}
\bibinfo{author}{\bibfnamefont{K.}~\bibnamefont{Doya}}, \emph{\bibinfo{title}{Bayesian {{Brain}}: {{Probabilistic Approaches}} to {{Neural Coding}}}} (\bibinfo{publisher}{MIT Press}, \bibinfo{year}{2007}), ISBN \bibinfo{isbn}{978-0-262-04238-3}.

\bibitem[{\citenamefont{Hills et~al.}(2015)\citenamefont{Hills, Todd, Lazer, Redish, and Couzin}}]{hillsExplorationExploitationSpace2015a}
\bibinfo{author}{\bibfnamefont{T.~T.} \bibnamefont{Hills}}, \bibinfo{author}{\bibfnamefont{P.~M.} \bibnamefont{Todd}}, \bibinfo{author}{\bibfnamefont{D.}~\bibnamefont{Lazer}}, \bibinfo{author}{\bibfnamefont{A.~D.} \bibnamefont{Redish}}, \bibnamefont{and} \bibinfo{author}{\bibfnamefont{I.~D.} \bibnamefont{Couzin}}, \bibinfo{journal}{Trends in Cognitive Sciences} \textbf{\bibinfo{volume}{19}}, \bibinfo{pages}{46} (\bibinfo{year}{2015}).

\bibitem[{\citenamefont{Friston et~al.}(2016)\citenamefont{Friston, FitzGerald, Rigoli, Schwartenbeck, O~Doherty, and Pezzulo}}]{fristonActiveInferenceLearning2016}
\bibinfo{author}{\bibfnamefont{K.}~\bibnamefont{Friston}}, \bibinfo{author}{\bibfnamefont{T.}~\bibnamefont{FitzGerald}}, \bibinfo{author}{\bibfnamefont{F.}~\bibnamefont{Rigoli}}, \bibinfo{author}{\bibfnamefont{P.}~\bibnamefont{Schwartenbeck}}, \bibinfo{author}{\bibfnamefont{J.}~\bibnamefont{O~Doherty}}, \bibnamefont{and} \bibinfo{author}{\bibfnamefont{G.}~\bibnamefont{Pezzulo}}, \bibinfo{journal}{Neurosci Biobehav Rev} \textbf{\bibinfo{volume}{68}}, \bibinfo{pages}{862} (\bibinfo{year}{2016}).

\bibitem[{\citenamefont{Tishby and Zaslavsky}(2015)}]{Tishby2015}
\bibinfo{author}{\bibfnamefont{N.}~\bibnamefont{Tishby}} \bibnamefont{and} \bibinfo{author}{\bibfnamefont{N.}~\bibnamefont{Zaslavsky}}, in \emph{\bibinfo{booktitle}{{{IEEE}} Information Theory Workshop ({{ITW}})}} (\bibinfo{year}{2015}).

\bibitem[{\citenamefont{Helias and Dahmen}(2020)}]{heliasStatisticalFieldTheory2020a}
\bibinfo{author}{\bibfnamefont{M.}~\bibnamefont{Helias}} \bibnamefont{and} \bibinfo{author}{\bibfnamefont{D.}~\bibnamefont{Dahmen}}, \emph{\bibinfo{title}{Statistical {{Field Theory}} for {{Neural Networks}}}}, vol. \bibinfo{volume}{970} of \emph{\bibinfo{series}{Lecture {{Notes}} in {{Physics}}}} (\bibinfo{publisher}{Springer International Publishing}, \bibinfo{address}{Cham}, \bibinfo{year}{2020}), ISBN \bibinfo{isbn}{978-3-030-46443-1 978-3-030-46444-8}.

\bibitem[{\citenamefont{Parr et~al.}(2022)\citenamefont{Parr, Pezzulo, and Friston}}]{parrActiveInferenceFree2022}
\bibinfo{author}{\bibfnamefont{T.}~\bibnamefont{Parr}}, \bibinfo{author}{\bibfnamefont{G.}~\bibnamefont{Pezzulo}}, \bibnamefont{and} \bibinfo{author}{\bibfnamefont{K.~J.} \bibnamefont{Friston}}, \emph{\bibinfo{title}{Active {{Inference}}: {{The Free Energy Principle}} in {{Mind}}, {{Brain}}, and {{Behavior}}}} (\bibinfo{publisher}{The MIT Press}, \bibinfo{year}{2022}), ISBN \bibinfo{isbn}{978-0-262-36997-8}.

\bibitem[{\citenamefont{Masson}(2013)}]{massonOlfactorySearchesLimited2013b}
\bibinfo{author}{\bibfnamefont{J.-B.} \bibnamefont{Masson}}, \bibinfo{journal}{Proceedings of the National Academy of Sciences} \textbf{\bibinfo{volume}{110}}, \bibinfo{pages}{11261} (\bibinfo{year}{2013}).

\bibitem[{\citenamefont{Reddy et~al.}(2022)\citenamefont{Reddy, Murthy, and Vergassola}}]{reddyOlfactorySensingNavigation2022}
\bibinfo{author}{\bibfnamefont{G.}~\bibnamefont{Reddy}}, \bibinfo{author}{\bibfnamefont{V.~N.} \bibnamefont{Murthy}}, \bibnamefont{and} \bibinfo{author}{\bibfnamefont{M.}~\bibnamefont{Vergassola}}, \bibinfo{journal}{Annual Review of Condensed Matter Physics} \textbf{\bibinfo{volume}{13}}, \bibinfo{pages}{191} (\bibinfo{year}{2022}).

\bibitem[{\citenamefont{Lattimore and Szepesv{\'a}ri}(2020)}]{lattimoreBanditAlgorithms2020}
\bibinfo{author}{\bibfnamefont{T.}~\bibnamefont{Lattimore}} \bibnamefont{and} \bibinfo{author}{\bibfnamefont{C.}~\bibnamefont{Szepesv{\'a}ri}}, \emph{\bibinfo{title}{Bandit {{Algorithms}}}} (\bibinfo{publisher}{Cambridge University Press}, \bibinfo{address}{Cambridge}, \bibinfo{year}{2020}), ISBN \bibinfo{isbn}{978-1-108-48682-8}.

\bibitem[{\citenamefont{Silver et~al.}(2016)\citenamefont{Silver, Huang, Maddison, Guez, Sifre, {van den Driessche}, Schrittwieser, Antonoglou, Panneershelvam, Lanctot et~al.}}]{silverMasteringGameGo2016}
\bibinfo{author}{\bibfnamefont{D.}~\bibnamefont{Silver}}, \bibinfo{author}{\bibfnamefont{A.}~\bibnamefont{Huang}}, \bibinfo{author}{\bibfnamefont{C.~J.} \bibnamefont{Maddison}}, \bibinfo{author}{\bibfnamefont{A.}~\bibnamefont{Guez}}, \bibinfo{author}{\bibfnamefont{L.}~\bibnamefont{Sifre}}, \bibinfo{author}{\bibfnamefont{G.}~\bibnamefont{{van den Driessche}}}, \bibinfo{author}{\bibfnamefont{J.}~\bibnamefont{Schrittwieser}}, \bibinfo{author}{\bibfnamefont{I.}~\bibnamefont{Antonoglou}}, \bibinfo{author}{\bibfnamefont{V.}~\bibnamefont{Panneershelvam}}, \bibinfo{author}{\bibfnamefont{M.}~\bibnamefont{Lanctot}}, \bibnamefont{et~al.}, \bibinfo{journal}{Nature} \textbf{\bibinfo{volume}{529}}, \bibinfo{pages}{484} (\bibinfo{year}{2016}).

\bibitem[{\citenamefont{Bouneffouf et~al.}(2020)\citenamefont{Bouneffouf, Rish, and Aggarwal}}]{bouneffoufSurveyApplicationsMultiArmed2020}
\bibinfo{author}{\bibfnamefont{D.}~\bibnamefont{Bouneffouf}}, \bibinfo{author}{\bibfnamefont{I.}~\bibnamefont{Rish}}, \bibnamefont{and} \bibinfo{author}{\bibfnamefont{C.}~\bibnamefont{Aggarwal}}, in \emph{\bibinfo{booktitle}{2020 {{IEEE Congress}} on {{Evolutionary Computation}} ({{CEC}})}} (\bibinfo{year}{2020}), pp. \bibinfo{pages}{1--8}.

\bibitem[{\citenamefont{Burtini et~al.}(2015)\citenamefont{Burtini, Loeppky, and Lawrence}}]{burtiniSurveyOnlineExperiment2015}
\bibinfo{author}{\bibfnamefont{G.}~\bibnamefont{Burtini}}, \bibinfo{author}{\bibfnamefont{J.}~\bibnamefont{Loeppky}}, \bibnamefont{and} \bibinfo{author}{\bibfnamefont{R.}~\bibnamefont{Lawrence}}, \emph{\bibinfo{title}{A {{Survey}} of {{Online Experiment Design}} with the {{Stochastic Multi-Armed Bandit}}}} (\bibinfo{year}{2015}), \eprint{1510.00757}.

\bibitem[{\citenamefont{Durand et~al.}(2018)\citenamefont{Durand, Achilleos, Iacovides, Strati, Mitsis, and Pineau}}]{durandContextualBanditsAdapting2018}
\bibinfo{author}{\bibfnamefont{A.}~\bibnamefont{Durand}}, \bibinfo{author}{\bibfnamefont{C.}~\bibnamefont{Achilleos}}, \bibinfo{author}{\bibfnamefont{D.}~\bibnamefont{Iacovides}}, \bibinfo{author}{\bibfnamefont{K.}~\bibnamefont{Strati}}, \bibinfo{author}{\bibfnamefont{G.~D.} \bibnamefont{Mitsis}}, \bibnamefont{and} \bibinfo{author}{\bibfnamefont{J.}~\bibnamefont{Pineau}}, in \emph{\bibinfo{booktitle}{Proceedings of the 3rd {{Machine Learning}} for {{Healthcare Conference}}}} (\bibinfo{publisher}{PMLR}, \bibinfo{year}{2018}), pp. \bibinfo{pages}{67--82}, ISSN \bibinfo{issn}{2640-3498}.

\bibitem[{\citenamefont{Gentile et~al.}(2014)\citenamefont{Gentile, Li, and Zappella}}]{gentileOnlineClusteringBandits2014}
\bibinfo{author}{\bibfnamefont{C.}~\bibnamefont{Gentile}}, \bibinfo{author}{\bibfnamefont{S.}~\bibnamefont{Li}}, \bibnamefont{and} \bibinfo{author}{\bibfnamefont{G.}~\bibnamefont{Zappella}}, in \emph{\bibinfo{booktitle}{Proceedings of the 31st {{International Conference}} on {{Machine Learning}}}} (\bibinfo{publisher}{PMLR}, \bibinfo{year}{2014}), pp. \bibinfo{pages}{757--765}, ISSN \bibinfo{issn}{1938-7228}.

\bibitem[{\citenamefont{Kuleshov and Precup}(2014)}]{kuleshovAlgorithmsMultiarmedBandit2014}
\bibinfo{author}{\bibfnamefont{V.}~\bibnamefont{Kuleshov}} \bibnamefont{and} \bibinfo{author}{\bibfnamefont{D.}~\bibnamefont{Precup}}, \emph{\bibinfo{title}{Algorithms for multi-armed bandit problems}} (\bibinfo{year}{2014}), \eprint{1402.6028}.

\bibitem[{\citenamefont{Mary et~al.}(2015)\citenamefont{Mary, Gaudel, and Preux}}]{maryBanditsRecommenderSystems2015}
\bibinfo{author}{\bibfnamefont{J.}~\bibnamefont{Mary}}, \bibinfo{author}{\bibfnamefont{R.}~\bibnamefont{Gaudel}}, \bibnamefont{and} \bibinfo{author}{\bibfnamefont{P.}~\bibnamefont{Preux}}, in \emph{\bibinfo{booktitle}{Machine {{Learning}}, {{Optimization}}, and {{Big Data}}}}, edited by \bibinfo{editor}{\bibfnamefont{P.}~\bibnamefont{Pardalos}}, \bibinfo{editor}{\bibfnamefont{M.}~\bibnamefont{Pavone}}, \bibinfo{editor}{\bibfnamefont{G.~M.} \bibnamefont{Farinella}}, \bibnamefont{and} \bibinfo{editor}{\bibfnamefont{V.}~\bibnamefont{Cutello}} (\bibinfo{publisher}{Springer International Publishing}, \bibinfo{address}{Cham}, \bibinfo{year}{2015}), Lecture {{Notes}} in {{Computer Science}}, pp. \bibinfo{pages}{325--336}, ISBN \bibinfo{isbn}{978-3-319-27926-8}.

\bibitem[{\citenamefont{Stewart et~al.}(2012)\citenamefont{Stewart, Bekolay, and Eliasmith}}]{stewartLearningSelectActions2012}
\bibinfo{author}{\bibfnamefont{T.~C.} \bibnamefont{Stewart}}, \bibinfo{author}{\bibfnamefont{T.}~\bibnamefont{Bekolay}}, \bibnamefont{and} \bibinfo{author}{\bibfnamefont{C.}~\bibnamefont{Eliasmith}}, \bibinfo{journal}{Front. Neurosci.} \textbf{\bibinfo{volume}{6}} (\bibinfo{year}{2012}).

\bibitem[{\citenamefont{Lai and Robbins}(1985)}]{laiAsymptoticallyEfficientAdaptive1985}
\bibinfo{author}{\bibfnamefont{T.~L.} \bibnamefont{Lai}} \bibnamefont{and} \bibinfo{author}{\bibfnamefont{H.}~\bibnamefont{Robbins}}, \bibinfo{journal}{Advances in Applied Mathematics} \textbf{\bibinfo{volume}{6}}, \bibinfo{pages}{4} (\bibinfo{year}{1985}).

\bibitem[{\citenamefont{Kaufmann et~al.}(2012)\citenamefont{Kaufmann, Korda, and Munos}}]{kaufmannThompsonSamplingAsymptotically2012}
\bibinfo{author}{\bibfnamefont{E.}~\bibnamefont{Kaufmann}}, \bibinfo{author}{\bibfnamefont{N.}~\bibnamefont{Korda}}, \bibnamefont{and} \bibinfo{author}{\bibfnamefont{R.}~\bibnamefont{Munos}}, in \emph{\bibinfo{booktitle}{Algorithmic {{Learning Theory}}}}, edited by \bibinfo{editor}{\bibfnamefont{N.~H.} \bibnamefont{Bshouty}}, \bibinfo{editor}{\bibfnamefont{G.}~\bibnamefont{Stoltz}}, \bibinfo{editor}{\bibfnamefont{N.}~\bibnamefont{Vayatis}}, \bibnamefont{and} \bibinfo{editor}{\bibfnamefont{T.}~\bibnamefont{Zeugmann}} (\bibinfo{publisher}{Springer}, \bibinfo{address}{Berlin, Heidelberg}, \bibinfo{year}{2012}), Lecture {{Notes}} in {{Computer Science}}, pp. \bibinfo{pages}{199--213}, ISBN \bibinfo{isbn}{978-3-642-34106-9}.

\bibitem[{\citenamefont{Auer}(2000)}]{auerUsingUpperConfidence2000}
\bibinfo{author}{\bibfnamefont{P.}~\bibnamefont{Auer}}, in \emph{\bibinfo{booktitle}{Proceedings 41st {{Annual Symposium}} on {{Foundations}} of {{Computer Science}}}} (\bibinfo{publisher}{IEEE Comput. Soc}, \bibinfo{address}{Redondo Beach, CA, USA}, \bibinfo{year}{2000}), pp. \bibinfo{pages}{270--279}, ISBN \bibinfo{isbn}{978-0-7695-0850-4}.

\bibitem[{\citenamefont{Bubeck et~al.}(2011)\citenamefont{Bubeck, Munos, and Stoltz}}]{bubeckPureExplorationFinitelyarmed2011a}
\bibinfo{author}{\bibfnamefont{S.}~\bibnamefont{Bubeck}}, \bibinfo{author}{\bibfnamefont{R.}~\bibnamefont{Munos}}, \bibnamefont{and} \bibinfo{author}{\bibfnamefont{G.}~\bibnamefont{Stoltz}}, \bibinfo{journal}{Theoretical Computer Science} \textbf{\bibinfo{volume}{412}}, \bibinfo{pages}{1832} (\bibinfo{year}{2011}).

\bibitem[{\citenamefont{Jamieson et~al.}(2014)\citenamefont{Jamieson, Malloy, Nowak, and Bubeck}}]{jamiesonLilUCBOptimal2014}
\bibinfo{author}{\bibfnamefont{K.}~\bibnamefont{Jamieson}}, \bibinfo{author}{\bibfnamefont{M.}~\bibnamefont{Malloy}}, \bibinfo{author}{\bibfnamefont{R.}~\bibnamefont{Nowak}}, \bibnamefont{and} \bibinfo{author}{\bibfnamefont{S.}~\bibnamefont{Bubeck}}, in \emph{\bibinfo{booktitle}{Proceedings of {{The}} 27th {{Conference}} on {{Learning Theory}}}} (\bibinfo{publisher}{PMLR}, \bibinfo{year}{2014}), pp. \bibinfo{pages}{423--439}, ISSN \bibinfo{issn}{1938-7228}.

\bibitem[{\citenamefont{Kaufmann and Kalyanakrishnan}(2013)}]{kaufmannInformationComplexityBandit2013}
\bibinfo{author}{\bibfnamefont{E.}~\bibnamefont{Kaufmann}} \bibnamefont{and} \bibinfo{author}{\bibfnamefont{S.}~\bibnamefont{Kalyanakrishnan}}, in \emph{\bibinfo{booktitle}{Proceedings of the 26th {{Annual Conference}} on {{Learning Theory}}}} (\bibinfo{publisher}{PMLR}, \bibinfo{year}{2013}), pp. \bibinfo{pages}{228--251}, ISSN \bibinfo{issn}{1938-7228}.

\bibitem[{\citenamefont{Kalyanakrishnan et~al.}(2012)\citenamefont{Kalyanakrishnan, Tewari, Auer, and Stone}}]{kalyanakrishnanPACSubsetSelection2012a}
\bibinfo{author}{\bibfnamefont{S.}~\bibnamefont{Kalyanakrishnan}}, \bibinfo{author}{\bibfnamefont{A.}~\bibnamefont{Tewari}}, \bibinfo{author}{\bibfnamefont{P.}~\bibnamefont{Auer}}, \bibnamefont{and} \bibinfo{author}{\bibfnamefont{P.}~\bibnamefont{Stone}}, \bibinfo{journal}{Proceedings of the 29th International Conference on Machine Learning, ICML 2012}  (\bibinfo{year}{2012}).

\bibitem[{\citenamefont{Degenne and Koolen}(2019)}]{degennePureExplorationMultiple2019}
\bibinfo{author}{\bibfnamefont{R.}~\bibnamefont{Degenne}} \bibnamefont{and} \bibinfo{author}{\bibfnamefont{W.~M.} \bibnamefont{Koolen}}, in \emph{\bibinfo{booktitle}{Advances in {{Neural Information Processing Systems}}}}, edited by \bibinfo{editor}{\bibfnamefont{H.}~\bibnamefont{Wallach}}, \bibinfo{editor}{\bibfnamefont{H.}~\bibnamefont{Larochelle}}, \bibinfo{editor}{\bibfnamefont{A.}~\bibnamefont{Beygelzimer}}, \bibinfo{editor}{\bibfnamefont{F.}~\bibnamefont{d'{\null} {Alch{\'e}-Buc}}}, \bibinfo{editor}{\bibfnamefont{E.}~\bibnamefont{Fox}}, \bibnamefont{and} \bibinfo{editor}{\bibfnamefont{R.}~\bibnamefont{Garnett}} (\bibinfo{publisher}{Curran Associates, Inc.}, \bibinfo{year}{2019}), vol.~\bibinfo{volume}{32}.

\bibitem[{\citenamefont{Li et~al.}(2010)\citenamefont{Li, Chu, Langford, and Schapire}}]{liContextualbanditApproachPersonalized2010}
\bibinfo{author}{\bibfnamefont{L.}~\bibnamefont{Li}}, \bibinfo{author}{\bibfnamefont{W.}~\bibnamefont{Chu}}, \bibinfo{author}{\bibfnamefont{J.}~\bibnamefont{Langford}}, \bibnamefont{and} \bibinfo{author}{\bibfnamefont{R.~E.} \bibnamefont{Schapire}}, in \emph{\bibinfo{booktitle}{Proceedings of the 19th International Conference on {{World}} Wide Web}} (\bibinfo{publisher}{Association for Computing Machinery}, \bibinfo{address}{New York, NY, USA}, \bibinfo{year}{2010}), {{WWW}} '10, pp. \bibinfo{pages}{661--670}, ISBN \bibinfo{isbn}{978-1-60558-799-8}.

\bibitem[{\citenamefont{Lin and Bouneffouf}(2022)}]{linOptimalEpidemicControl2022}
\bibinfo{author}{\bibfnamefont{B.}~\bibnamefont{Lin}} \bibnamefont{and} \bibinfo{author}{\bibfnamefont{D.}~\bibnamefont{Bouneffouf}}, in \emph{\bibinfo{booktitle}{2022 {{IEEE International Conference}} on {{Fuzzy Systems}} ({{FUZZ-IEEE}})}} (\bibinfo{year}{2022}), pp. \bibinfo{pages}{1--8}, ISSN \bibinfo{issn}{1558-4739}.

\bibitem[{\citenamefont{Zhou}(2016)}]{zhouSurveyContextualMultiarmed2016}
\bibinfo{author}{\bibfnamefont{L.}~\bibnamefont{Zhou}}, \emph{\bibinfo{title}{A {{Survey}} on {{Contextual Multi-armed Bandits}}}} (\bibinfo{year}{2016}), \eprint{1508.03326}.

\bibitem[{\citenamefont{Auer}(2002)}]{auerUsingConfidenceBounds2002}
\bibinfo{author}{\bibfnamefont{P.}~\bibnamefont{Auer}}, \bibinfo{journal}{Journal of Machine Learning Research} \textbf{\bibinfo{volume}{3}}, \bibinfo{pages}{397} (\bibinfo{year}{2002}).

\bibitem[{\citenamefont{Lattimore and Szepesvari}(2017)}]{lattimoreEndOptimismAsymptotic2017}
\bibinfo{author}{\bibfnamefont{T.}~\bibnamefont{Lattimore}} \bibnamefont{and} \bibinfo{author}{\bibfnamefont{C.}~\bibnamefont{Szepesvari}}, in \emph{\bibinfo{booktitle}{Proceedings of the 20th {{International Conference}} on {{Artificial Intelligence}} and {{Statistics}}}} (\bibinfo{publisher}{PMLR}, \bibinfo{year}{2017}), pp. \bibinfo{pages}{728--737}, ISSN \bibinfo{issn}{2640-3498}.

\bibitem[{\citenamefont{Agrawal and Goyal}(2013)}]{agrawalThompsonSamplingContextual2013}
\bibinfo{author}{\bibfnamefont{S.}~\bibnamefont{Agrawal}} \bibnamefont{and} \bibinfo{author}{\bibfnamefont{N.}~\bibnamefont{Goyal}}, in \emph{\bibinfo{booktitle}{Proceedings of the 30th {{International Conference}} on {{Machine Learning}}}} (\bibinfo{publisher}{PMLR}, \bibinfo{year}{2013}), pp. \bibinfo{pages}{127--135}, ISSN \bibinfo{issn}{1938-7228}.

\bibitem[{\citenamefont{Villar et~al.}(2015)\citenamefont{Villar, Bowden, and Wason}}]{villarMultiarmedBanditModels2015}
\bibinfo{author}{\bibfnamefont{S.~S.} \bibnamefont{Villar}}, \bibinfo{author}{\bibfnamefont{J.}~\bibnamefont{Bowden}}, \bibnamefont{and} \bibinfo{author}{\bibfnamefont{J.}~\bibnamefont{Wason}}, \bibinfo{journal}{Statistical Science} \textbf{\bibinfo{volume}{30}}, \bibinfo{pages}{199} (\bibinfo{year}{2015}).

\bibitem[{\citenamefont{Villar}(2018)}]{villarBanditsStrategiesEvaluated2018}
\bibinfo{author}{\bibfnamefont{S.~S.} \bibnamefont{Villar}}, \bibinfo{journal}{Probability in the Engineering and Informational Sciences} \textbf{\bibinfo{volume}{32}}, \bibinfo{pages}{229} (\bibinfo{year}{2018}).

\bibitem[{\citenamefont{{de Heide} et~al.}(2024)\citenamefont{{de Heide}, Cheshire, M{\'e}nard, and Carpentier}}]{deheideBanditsManyOptimal2024}
\bibinfo{author}{\bibfnamefont{R.}~\bibnamefont{{de Heide}}}, \bibinfo{author}{\bibfnamefont{J.}~\bibnamefont{Cheshire}}, \bibinfo{author}{\bibfnamefont{P.}~\bibnamefont{M{\'e}nard}}, \bibnamefont{and} \bibinfo{author}{\bibfnamefont{A.}~\bibnamefont{Carpentier}}, in \emph{\bibinfo{booktitle}{Proceedings of the 35th {{International Conference}} on {{Neural Information Processing Systems}}}} (\bibinfo{publisher}{Curran Associates Inc.}, \bibinfo{address}{Red Hook, NY, USA}, \bibinfo{year}{2024}), {{NIPS}} '21, pp. \bibinfo{pages}{22457--22469}, ISBN \bibinfo{isbn}{978-1-7138-4539-3}.

\bibitem[{\citenamefont{Bonald and Proutiere}(2013)}]{bonaldTwoTargetAlgorithmsInfiniteArmed2013}
\bibinfo{author}{\bibfnamefont{T.}~\bibnamefont{Bonald}} \bibnamefont{and} \bibinfo{author}{\bibfnamefont{A.}~\bibnamefont{Proutiere}}, in \emph{\bibinfo{booktitle}{Advances in {{Neural Information Processing Systems}}}}, edited by \bibinfo{editor}{\bibfnamefont{C.~J.} \bibnamefont{Burges}}, \bibinfo{editor}{\bibfnamefont{L.}~\bibnamefont{Bottou}}, \bibinfo{editor}{\bibfnamefont{M.}~\bibnamefont{Welling}}, \bibinfo{editor}{\bibfnamefont{Z.}~\bibnamefont{Ghahramani}}, \bibnamefont{and} \bibinfo{editor}{\bibfnamefont{K.~Q.} \bibnamefont{Weinberger}} (\bibinfo{publisher}{Curran Associates, Inc.}, \bibinfo{year}{2013}), vol.~\bibinfo{volume}{26}.

\bibitem[{\citenamefont{Berry et~al.}(1997)\citenamefont{Berry, Chen, Zame, Heath, and Shepp}}]{berryBanditProblemsInfinitely1997}
\bibinfo{author}{\bibfnamefont{D.~A.} \bibnamefont{Berry}}, \bibinfo{author}{\bibfnamefont{R.~W.} \bibnamefont{Chen}}, \bibinfo{author}{\bibfnamefont{A.}~\bibnamefont{Zame}}, \bibinfo{author}{\bibfnamefont{D.~C.} \bibnamefont{Heath}}, \bibnamefont{and} \bibinfo{author}{\bibfnamefont{L.~A.} \bibnamefont{Shepp}}, \bibinfo{journal}{The Annals of Statistics} \textbf{\bibinfo{volume}{25}}, \bibinfo{pages}{2103} (\bibinfo{year}{1997}).

\bibitem[{\citenamefont{Wang et~al.}(2008)\citenamefont{Wang, Audibert, and Munos}}]{wangAlgorithmsInfinitelyManyArmed2008}
\bibinfo{author}{\bibfnamefont{Y.}~\bibnamefont{Wang}}, \bibinfo{author}{\bibfnamefont{J.-y.} \bibnamefont{Audibert}}, \bibnamefont{and} \bibinfo{author}{\bibfnamefont{R.}~\bibnamefont{Munos}}, in \emph{\bibinfo{booktitle}{Advances in {{Neural Information Processing Systems}}}} (\bibinfo{publisher}{Curran Associates, Inc.}, \bibinfo{year}{2008}), vol.~\bibinfo{volume}{21}.

\bibitem[{\citenamefont{Baudry et~al.}(2020)\citenamefont{Baudry, Kaufmann, and Maillard}}]{baudrySubsamplingEfficientNonparametric2020}
\bibinfo{author}{\bibfnamefont{D.}~\bibnamefont{Baudry}}, \bibinfo{author}{\bibfnamefont{E.}~\bibnamefont{Kaufmann}}, \bibnamefont{and} \bibinfo{author}{\bibfnamefont{O.-A.} \bibnamefont{Maillard}}, in \emph{\bibinfo{booktitle}{Proceedings of the 34th {{International Conference}} on {{Neural Information Processing Systems}}}} (\bibinfo{publisher}{Curran Associates Inc.}, \bibinfo{address}{Red Hook, NY, USA}, \bibinfo{year}{2020}), {{NIPS}}'20, pp. \bibinfo{pages}{5468--5478}, ISBN \bibinfo{isbn}{978-1-7138-2954-6}.

\bibitem[{\citenamefont{Guha et~al.}(2010)\citenamefont{Guha, Munagala, and Shi}}]{guhaApproximationAlgorithmsRestless2010}
\bibinfo{author}{\bibfnamefont{S.}~\bibnamefont{Guha}}, \bibinfo{author}{\bibfnamefont{K.}~\bibnamefont{Munagala}}, \bibnamefont{and} \bibinfo{author}{\bibfnamefont{P.}~\bibnamefont{Shi}}, \bibinfo{journal}{J. ACM} \textbf{\bibinfo{volume}{58}}, \bibinfo{pages}{3:1} (\bibinfo{year}{2010}).

\bibitem[{\citenamefont{Levine et~al.}(2017)\citenamefont{Levine, Crammer, and Mannor}}]{levineRottingBandits2017}
\bibinfo{author}{\bibfnamefont{N.}~\bibnamefont{Levine}}, \bibinfo{author}{\bibfnamefont{K.}~\bibnamefont{Crammer}}, \bibnamefont{and} \bibinfo{author}{\bibfnamefont{S.}~\bibnamefont{Mannor}}, in \emph{\bibinfo{booktitle}{Advances in {{Neural Information Processing Systems}}}}, edited by \bibinfo{editor}{\bibfnamefont{I.}~\bibnamefont{Guyon}}, \bibinfo{editor}{\bibfnamefont{U.~V.} \bibnamefont{Luxburg}}, \bibinfo{editor}{\bibfnamefont{S.}~\bibnamefont{Bengio}}, \bibinfo{editor}{\bibfnamefont{H.}~\bibnamefont{Wallach}}, \bibinfo{editor}{\bibfnamefont{R.}~\bibnamefont{Fergus}}, \bibinfo{editor}{\bibfnamefont{S.}~\bibnamefont{Vishwanathan}}, \bibnamefont{and} \bibinfo{editor}{\bibfnamefont{R.}~\bibnamefont{Garnett}} (\bibinfo{publisher}{Curran Associates, Inc.}, \bibinfo{year}{2017}), vol.~\bibinfo{volume}{30}, pp. \bibinfo{pages}{3074--3083}.

\bibitem[{\citenamefont{Auer et~al.}(2002)\citenamefont{Auer, {Cesa-Bianchi}, and Fischer}}]{auerFinitetimeAnalysisMultiarmed2002}
\bibinfo{author}{\bibfnamefont{P.}~\bibnamefont{Auer}}, \bibinfo{author}{\bibfnamefont{N.}~\bibnamefont{{Cesa-Bianchi}}}, \bibnamefont{and} \bibinfo{author}{\bibfnamefont{P.}~\bibnamefont{Fischer}}, \bibinfo{journal}{Machine Learning} \textbf{\bibinfo{volume}{47}}, \bibinfo{pages}{235} (\bibinfo{year}{2002}).

\bibitem[{\citenamefont{Bastani et~al.}(2021)\citenamefont{Bastani, Bayati, and Khosravi}}]{bastaniMostlyExplorationFreeAlgorithms2021}
\bibinfo{author}{\bibfnamefont{H.}~\bibnamefont{Bastani}}, \bibinfo{author}{\bibfnamefont{M.}~\bibnamefont{Bayati}}, \bibnamefont{and} \bibinfo{author}{\bibfnamefont{K.}~\bibnamefont{Khosravi}}, \bibinfo{journal}{Management Science} \textbf{\bibinfo{volume}{67}}, \bibinfo{pages}{1329} (\bibinfo{year}{2021}).

\bibitem[{\citenamefont{M{\'e}nard and Garivier}(2017)}]{menardMinimaxAsymptoticallyOptimal2017}
\bibinfo{author}{\bibfnamefont{P.}~\bibnamefont{M{\'e}nard}} \bibnamefont{and} \bibinfo{author}{\bibfnamefont{A.}~\bibnamefont{Garivier}}, in \emph{\bibinfo{booktitle}{Proceedings of the 28th {{International Conference}} on {{Algorithmic Learning Theory}}}} (\bibinfo{publisher}{PMLR}, \bibinfo{year}{2017}), pp. \bibinfo{pages}{223--237}, ISSN \bibinfo{issn}{2640-3498}.

\bibitem[{\citenamefont{Russo and Roy}(2016)}]{russoInformationTheoreticAnalysisThompson2016}
\bibinfo{author}{\bibfnamefont{D.}~\bibnamefont{Russo}} \bibnamefont{and} \bibinfo{author}{\bibfnamefont{B.~V.} \bibnamefont{Roy}}, \bibinfo{journal}{Journal of Machine Learning Research} \textbf{\bibinfo{volume}{17}}, \bibinfo{pages}{1} (\bibinfo{year}{2016}).

\bibitem[{\citenamefont{Reddy et~al.}(2016)\citenamefont{Reddy, Celani, and Vergassola}}]{reddyInfomaxStrategiesOptimal2016}
\bibinfo{author}{\bibfnamefont{G.}~\bibnamefont{Reddy}}, \bibinfo{author}{\bibfnamefont{A.}~\bibnamefont{Celani}}, \bibnamefont{and} \bibinfo{author}{\bibfnamefont{M.}~\bibnamefont{Vergassola}}, \bibinfo{journal}{Journal of Statistical Physics} \textbf{\bibinfo{volume}{163}}, \bibinfo{pages}{1454} (\bibinfo{year}{2016}).

\bibitem[{\citenamefont{Hung et~al.}(2020)\citenamefont{Hung, Hsieh, Liu, and Kumar}}]{hungRewardBiasedMaximumLikelihood2020}
\bibinfo{author}{\bibfnamefont{Y.-H.} \bibnamefont{Hung}}, \bibinfo{author}{\bibfnamefont{P.-C.} \bibnamefont{Hsieh}}, \bibinfo{author}{\bibfnamefont{X.}~\bibnamefont{Liu}}, \bibnamefont{and} \bibinfo{author}{\bibfnamefont{P.~R.} \bibnamefont{Kumar}}, \emph{\bibinfo{title}{Reward-{{Biased Maximum Likelihood Estimation}} for {{Linear Stochastic Bandits}}}} (\bibinfo{year}{2020}), \eprint{2010.04091}.

\bibitem[{\citenamefont{Tirinzoni et~al.}(2020)\citenamefont{Tirinzoni, Pirotta, Restelli, and Lazaric}}]{tirinzoniAsymptoticallyOptimalPrimaldual2020}
\bibinfo{author}{\bibfnamefont{A.}~\bibnamefont{Tirinzoni}}, \bibinfo{author}{\bibfnamefont{M.}~\bibnamefont{Pirotta}}, \bibinfo{author}{\bibfnamefont{M.}~\bibnamefont{Restelli}}, \bibnamefont{and} \bibinfo{author}{\bibfnamefont{A.}~\bibnamefont{Lazaric}}, in \emph{\bibinfo{booktitle}{Proceedings of the 34th {{International Conference}} on {{Neural Information Processing Systems}}}} (\bibinfo{publisher}{Curran Associates Inc.}, \bibinfo{address}{Red Hook, NY, USA}, \bibinfo{year}{2020}), {{NIPS}} '20, pp. \bibinfo{pages}{1417--1427}, ISBN \bibinfo{isbn}{978-1-7138-2954-6}.

\bibitem[{\citenamefont{{Barbier-Chebbah} et~al.}(2023{\natexlab{a}})\citenamefont{{Barbier-Chebbah}, Vestergaard, and Masson}}]{barbier-chebbahApproximateInformationEfficient2023}
\bibinfo{author}{\bibfnamefont{A.}~\bibnamefont{{Barbier-Chebbah}}}, \bibinfo{author}{\bibfnamefont{C.~L.} \bibnamefont{Vestergaard}}, \bibnamefont{and} \bibinfo{author}{\bibfnamefont{J.-B.} \bibnamefont{Masson}}, \emph{\bibinfo{title}{Approximate information for efficient exploration-exploitation strategies}} (\bibinfo{year}{2023}{\natexlab{a}}), \eprint{2307.01563}.

\bibitem[{\citenamefont{{Barbier-Chebbah} et~al.}(2023{\natexlab{b}})\citenamefont{{Barbier-Chebbah}, Vestergaard, Masson, and Boursier}}]{barbier-chebbahApproximateInformationMaximization2023}
\bibinfo{author}{\bibfnamefont{A.}~\bibnamefont{{Barbier-Chebbah}}}, \bibinfo{author}{\bibfnamefont{C.~L.} \bibnamefont{Vestergaard}}, \bibinfo{author}{\bibfnamefont{J.-B.} \bibnamefont{Masson}}, \bibnamefont{and} \bibinfo{author}{\bibfnamefont{E.}~\bibnamefont{Boursier}}, \emph{\bibinfo{title}{Approximate information maximization for bandit games}} (\bibinfo{year}{2023}{\natexlab{b}}), \eprint{2310.12563}.

\bibitem[{\citenamefont{Kaufmann et~al.}(2016)\citenamefont{Kaufmann, Capp{\'e}, and Garivier}}]{kaufmannComplexityBestarmIdentification2016}
\bibinfo{author}{\bibfnamefont{E.}~\bibnamefont{Kaufmann}}, \bibinfo{author}{\bibfnamefont{O.}~\bibnamefont{Capp{\'e}}}, \bibnamefont{and} \bibinfo{author}{\bibfnamefont{A.}~\bibnamefont{Garivier}}, \bibinfo{journal}{J. Mach. Learn. Res.} \textbf{\bibinfo{volume}{17}}, \bibinfo{pages}{1} (\bibinfo{year}{2016}).

\bibitem[{\citenamefont{Kalyanakrishnan and Stone}(2010)}]{kalyanakrishnanEfficientSelectionMultiple2010}
\bibinfo{author}{\bibfnamefont{S.}~\bibnamefont{Kalyanakrishnan}} \bibnamefont{and} \bibinfo{author}{\bibfnamefont{P.}~\bibnamefont{Stone}}, in \emph{\bibinfo{booktitle}{Proceedings of the 27th {{International Conference}} on {{International Conference}} on {{Machine Learning}}}} (\bibinfo{publisher}{Omnipress}, \bibinfo{address}{Madison, WI, USA}, \bibinfo{year}{2010}), {{ICML}}'10, pp. \bibinfo{pages}{511--518}, ISBN \bibinfo{isbn}{978-1-60558-907-7}.

\bibitem[{\citenamefont{Mannor and Tsitsiklis}(2004)}]{mannorSampleComplexityExploration2004}
\bibinfo{author}{\bibfnamefont{S.}~\bibnamefont{Mannor}} \bibnamefont{and} \bibinfo{author}{\bibfnamefont{J.~N.} \bibnamefont{Tsitsiklis}}, \bibinfo{journal}{Journal of Machine Learning Research} \textbf{\bibinfo{volume}{5}}, \bibinfo{pages}{623} (\bibinfo{year}{2004}).

\bibitem[{\citenamefont{Zhou et~al.}(2014)\citenamefont{Zhou, Chen, and Li}}]{zhouOptimalPACMultiple2014}
\bibinfo{author}{\bibfnamefont{Y.}~\bibnamefont{Zhou}}, \bibinfo{author}{\bibfnamefont{X.}~\bibnamefont{Chen}}, \bibnamefont{and} \bibinfo{author}{\bibfnamefont{J.}~\bibnamefont{Li}}, in \emph{\bibinfo{booktitle}{Proceedings of the 31st {{International Conference}} on {{Machine Learning}}}} (\bibinfo{publisher}{PMLR}, \bibinfo{year}{2014}), pp. \bibinfo{pages}{217--225}, ISSN \bibinfo{issn}{1938-7228}.

\bibitem[{\citenamefont{Bubeck et~al.}(2013)\citenamefont{Bubeck, Wang, and Viswanathan}}]{bubeckMultipleIdentificationsMultiArmed2013}
\bibinfo{author}{\bibfnamefont{S.}~\bibnamefont{Bubeck}}, \bibinfo{author}{\bibfnamefont{T.}~\bibnamefont{Wang}}, \bibnamefont{and} \bibinfo{author}{\bibfnamefont{N.}~\bibnamefont{Viswanathan}}, in \emph{\bibinfo{booktitle}{Proceedings of the 30th {{International Conference}} on {{Machine Learning}}}} (\bibinfo{publisher}{PMLR}, \bibinfo{year}{2013}), pp. \bibinfo{pages}{258--265}, ISSN \bibinfo{issn}{1938-7228}.

\bibitem[{\citenamefont{Gabillon et~al.}(2012)\citenamefont{Gabillon, Ghavamzadeh, and Lazaric}}]{gabillonBestArmIdentification2012}
\bibinfo{author}{\bibfnamefont{V.}~\bibnamefont{Gabillon}}, \bibinfo{author}{\bibfnamefont{M.}~\bibnamefont{Ghavamzadeh}}, \bibnamefont{and} \bibinfo{author}{\bibfnamefont{A.}~\bibnamefont{Lazaric}}, in \emph{\bibinfo{booktitle}{Advances in {{Neural Information Processing Systems}}}}, edited by \bibinfo{editor}{\bibfnamefont{F.}~\bibnamefont{Pereira}}, \bibinfo{editor}{\bibfnamefont{C.~J.} \bibnamefont{Burges}}, \bibinfo{editor}{\bibfnamefont{L.}~\bibnamefont{Bottou}}, \bibnamefont{and} \bibinfo{editor}{\bibfnamefont{K.~Q.} \bibnamefont{Weinberger}} (\bibinfo{publisher}{Curran Associates, Inc.}, \bibinfo{year}{2012}), vol.~\bibinfo{volume}{25}.

\bibitem[{\citenamefont{Audibert and Bubeck}(2010)}]{audibertBestArmIdentification2010}
\bibinfo{author}{\bibfnamefont{J.-Y.} \bibnamefont{Audibert}} \bibnamefont{and} \bibinfo{author}{\bibfnamefont{S.}~\bibnamefont{Bubeck}}, in \emph{\bibinfo{booktitle}{{{COLT}} - 23th {{Conference}} on {{Learning Theory}} - 2010}} (\bibinfo{year}{2010}), p. \bibinfo{pages}{13 p.}

\bibitem[{\citenamefont{Majumdar et~al.}(2020)\citenamefont{Majumdar, Pal, and Schehr}}]{majumdarExtremeValueStatistics2020a}
\bibinfo{author}{\bibfnamefont{S.~N.} \bibnamefont{Majumdar}}, \bibinfo{author}{\bibfnamefont{A.}~\bibnamefont{Pal}}, \bibnamefont{and} \bibinfo{author}{\bibfnamefont{G.}~\bibnamefont{Schehr}}, \bibinfo{journal}{Physics Reports} \textbf{\bibinfo{volume}{840}}, \bibinfo{pages}{1} (\bibinfo{year}{2020}).

\bibitem[{\citenamefont{Carpentier and Valko}(2015)}]{carpentierSimpleRegretInfinitely2015}
\bibinfo{author}{\bibfnamefont{A.}~\bibnamefont{Carpentier}} \bibnamefont{and} \bibinfo{author}{\bibfnamefont{M.}~\bibnamefont{Valko}}, in \emph{\bibinfo{booktitle}{Proceedings of the 32nd {{International Conference}} on {{Machine Learning}}}} (\bibinfo{publisher}{PMLR}, \bibinfo{year}{2015}), pp. \bibinfo{pages}{1133--1141}, ISSN \bibinfo{issn}{1938-7228}.

\bibitem[{\citenamefont{Rejwan and Mansour}(2020)}]{rejwanTop$k$CombinatorialBandits2020}
\bibinfo{author}{\bibfnamefont{I.}~\bibnamefont{Rejwan}} \bibnamefont{and} \bibinfo{author}{\bibfnamefont{Y.}~\bibnamefont{Mansour}}, in \emph{\bibinfo{booktitle}{Proceedings of the 31st {{International Conference}} on {{Algorithmic Learning Theory}}}} (\bibinfo{publisher}{PMLR}, \bibinfo{year}{2020}), pp. \bibinfo{pages}{752--776}, ISSN \bibinfo{issn}{2640-3498}.

\bibitem[{\citenamefont{Katariya et~al.}(2019)\citenamefont{Katariya, Tripathy, and Nowak}}]{katariyaMaxGapBanditAdaptive2019}
\bibinfo{author}{\bibfnamefont{S.}~\bibnamefont{Katariya}}, \bibinfo{author}{\bibfnamefont{A.}~\bibnamefont{Tripathy}}, \bibnamefont{and} \bibinfo{author}{\bibfnamefont{R.}~\bibnamefont{Nowak}}, in \emph{\bibinfo{booktitle}{Proceedings of the 33rd {{International Conference}} on {{Neural Information Processing Systems}}}} (\bibinfo{publisher}{Curran Associates Inc.}, \bibinfo{address}{Red Hook, NY, USA}, \bibinfo{year}{2019}), \bibinfo{number}{991}, pp. \bibinfo{pages}{11047--11057}.

\bibitem[{\citenamefont{Locatelli et~al.}(2016)\citenamefont{Locatelli, Gutzeit, and Carpentier}}]{locatelliOptimalAlgorithmThresholding2016}
\bibinfo{author}{\bibfnamefont{A.}~\bibnamefont{Locatelli}}, \bibinfo{author}{\bibfnamefont{M.}~\bibnamefont{Gutzeit}}, \bibnamefont{and} \bibinfo{author}{\bibfnamefont{A.}~\bibnamefont{Carpentier}}, in \emph{\bibinfo{booktitle}{Proceedings of {{The}} 33rd {{International Conference}} on {{Machine Learning}}}} (\bibinfo{publisher}{PMLR}, \bibinfo{year}{2016}), pp. \bibinfo{pages}{1690--1698}, ISSN \bibinfo{issn}{1938-7228}.

\bibitem[{\citenamefont{Mason et~al.}(2020)\citenamefont{Mason, Jain, Tripathy, and Nowak}}]{masonFindingAllTextbackslash2020}
\bibinfo{author}{\bibfnamefont{B.}~\bibnamefont{Mason}}, \bibinfo{author}{\bibfnamefont{L.}~\bibnamefont{Jain}}, \bibinfo{author}{\bibfnamefont{A.}~\bibnamefont{Tripathy}}, \bibnamefont{and} \bibinfo{author}{\bibfnamefont{R.}~\bibnamefont{Nowak}}, in \emph{\bibinfo{booktitle}{Advances in {{Neural Information Processing Systems}}}}, edited by \bibinfo{editor}{\bibfnamefont{H.}~\bibnamefont{Larochelle}}, \bibinfo{editor}{\bibfnamefont{M.}~\bibnamefont{Ranzato}}, \bibinfo{editor}{\bibfnamefont{R.}~\bibnamefont{Hadsell}}, \bibinfo{editor}{\bibfnamefont{M.~F.} \bibnamefont{Balcan}}, \bibnamefont{and} \bibinfo{editor}{\bibfnamefont{H.}~\bibnamefont{Lin}} (\bibinfo{publisher}{Curran Associates, Inc.}, \bibinfo{year}{2020}), vol.~\bibinfo{volume}{33}, pp. \bibinfo{pages}{20707--20718}.

\bibitem[{\citenamefont{Ren et~al.}(2019)\citenamefont{Ren, Liu, and Shroff}}]{renExploring$k$Out2019}
\bibinfo{author}{\bibfnamefont{W.}~\bibnamefont{Ren}}, \bibinfo{author}{\bibfnamefont{J.}~\bibnamefont{Liu}}, \bibnamefont{and} \bibinfo{author}{\bibfnamefont{N.~B.} \bibnamefont{Shroff}}, in \emph{\bibinfo{booktitle}{Proceedings of the {{Twenty-Second International Conference}} on {{Artificial Intelligence}} and {{Statistics}}}} (\bibinfo{publisher}{PMLR}, \bibinfo{year}{2019}), pp. \bibinfo{pages}{2820--2828}, ISSN \bibinfo{issn}{2640-3498}.

\bibitem[{\citenamefont{Zhuang et~al.}(2017)\citenamefont{Zhuang, Wang, and Wang}}]{zhuangIdentifyingOutlierArms2017}
\bibinfo{author}{\bibfnamefont{H.}~\bibnamefont{Zhuang}}, \bibinfo{author}{\bibfnamefont{C.}~\bibnamefont{Wang}}, \bibnamefont{and} \bibinfo{author}{\bibfnamefont{Y.}~\bibnamefont{Wang}}, in \emph{\bibinfo{booktitle}{Advances in {{Neural Information Processing Systems}}}}, edited by \bibinfo{editor}{\bibfnamefont{I.}~\bibnamefont{Guyon}}, \bibinfo{editor}{\bibfnamefont{U.~V.} \bibnamefont{Luxburg}}, \bibinfo{editor}{\bibfnamefont{S.}~\bibnamefont{Bengio}}, \bibinfo{editor}{\bibfnamefont{H.}~\bibnamefont{Wallach}}, \bibinfo{editor}{\bibfnamefont{R.}~\bibnamefont{Fergus}}, \bibinfo{editor}{\bibfnamefont{S.}~\bibnamefont{Vishwanathan}}, \bibnamefont{and} \bibinfo{editor}{\bibfnamefont{R.}~\bibnamefont{Garnett}} (\bibinfo{publisher}{Curran Associates, Inc.}, \bibinfo{year}{2017}), vol.~\bibinfo{volume}{30}.

\bibitem[{\citenamefont{Abe et~al.}(2003)\citenamefont{Abe, Biermann, and Long}}]{abeReinforcementLearningImmediate2003}
\bibinfo{author}{\bibfnamefont{N.}~\bibnamefont{Abe}}, \bibinfo{author}{\bibfnamefont{A.~W.} \bibnamefont{Biermann}}, \bibnamefont{and} \bibinfo{author}{\bibfnamefont{P.~M.} \bibnamefont{Long}}, \bibinfo{journal}{Algorithmica} \textbf{\bibinfo{volume}{37}}, \bibinfo{pages}{263} (\bibinfo{year}{2003}).

\bibitem[{\citenamefont{Chu et~al.}(2011)\citenamefont{Chu, Li, Reyzin, and Schapire}}]{chuContextualBanditsLinear2011}
\bibinfo{author}{\bibfnamefont{W.}~\bibnamefont{Chu}}, \bibinfo{author}{\bibfnamefont{L.}~\bibnamefont{Li}}, \bibinfo{author}{\bibfnamefont{L.}~\bibnamefont{Reyzin}}, \bibnamefont{and} \bibinfo{author}{\bibfnamefont{R.}~\bibnamefont{Schapire}}, in \emph{\bibinfo{booktitle}{Proceedings of the {{Fourteenth International Conference}} on {{Artificial Intelligence}} and {{Statistics}}}} (\bibinfo{publisher}{{JMLR Workshop and Conference Proceedings}}, \bibinfo{year}{2011}), pp. \bibinfo{pages}{208--214}, ISSN \bibinfo{issn}{1938-7228}.

\bibitem[{\citenamefont{Slivkins}(2011)}]{slivkinsContextualBanditsSimilarity2011}
\bibinfo{author}{\bibfnamefont{A.}~\bibnamefont{Slivkins}}, in \emph{\bibinfo{booktitle}{Proceedings of the 24th {{Annual Conference}} on {{Learning Theory}}}} (\bibinfo{publisher}{{JMLR Workshop and Conference Proceedings}}, \bibinfo{year}{2011}), pp. \bibinfo{pages}{679--702}, ISSN \bibinfo{issn}{1938-7228}.

\bibitem[{\citenamefont{Combes et~al.}(2017)\citenamefont{Combes, Magureanu, and Proutiere}}]{combesMinimalExplorationStructured2017a}
\bibinfo{author}{\bibfnamefont{R.}~\bibnamefont{Combes}}, \bibinfo{author}{\bibfnamefont{S.}~\bibnamefont{Magureanu}}, \bibnamefont{and} \bibinfo{author}{\bibfnamefont{A.}~\bibnamefont{Proutiere}}, in \emph{\bibinfo{booktitle}{Proceedings of the 31st {{International Conference}} on {{Neural Information Processing Systems}}}} (\bibinfo{publisher}{Curran Associates Inc.}, \bibinfo{address}{Red Hook, NY, USA}, \bibinfo{year}{2017}), {{NIPS}}'17, pp. \bibinfo{pages}{1761--1769}, ISBN \bibinfo{isbn}{978-1-5108-6096-4}.

\bibitem[{\citenamefont{Dani et~al.}(2008)\citenamefont{Dani, Hayes, and Kakade}}]{daniStochasticLinearOptimization2008a}
\bibinfo{author}{\bibfnamefont{V.}~\bibnamefont{Dani}}, \bibinfo{author}{\bibfnamefont{T.~P.} \bibnamefont{Hayes}}, \bibnamefont{and} \bibinfo{author}{\bibfnamefont{S.~M.} \bibnamefont{Kakade}}, in \emph{\bibinfo{booktitle}{21st {{Annual Conference}} on {{Learning Theory}} - {{COLT}} 2008, {{Helsinki}}, {{Finland}}, {{July}} 9-12, 2008}}, edited by \bibinfo{editor}{\bibfnamefont{R.~A.} \bibnamefont{Servedio}} \bibnamefont{and} \bibinfo{editor}{\bibfnamefont{T.}~\bibnamefont{Zhang}} (\bibinfo{publisher}{Omnipress}, \bibinfo{year}{2008}), pp. \bibinfo{pages}{355--366}.

\bibitem[{\citenamefont{{Abbasi-yadkori} et~al.}(2011)\citenamefont{{Abbasi-yadkori}, P{\'a}l, and Szepesv{\'a}ri}}]{abbasi-yadkoriImprovedAlgorithmsLinear2011}
\bibinfo{author}{\bibfnamefont{Y.}~\bibnamefont{{Abbasi-yadkori}}}, \bibinfo{author}{\bibfnamefont{D.}~\bibnamefont{P{\'a}l}}, \bibnamefont{and} \bibinfo{author}{\bibfnamefont{C.}~\bibnamefont{Szepesv{\'a}ri}}, in \emph{\bibinfo{booktitle}{Advances in {{Neural Information Processing Systems}}}} (\bibinfo{publisher}{Curran Associates, Inc.}, \bibinfo{year}{2011}), vol.~\bibinfo{volume}{24}.

\bibitem[{\citenamefont{Hao et~al.}(2020)\citenamefont{Hao, Lattimore, and Szepesvari}}]{haoAdaptiveExplorationLinear2020}
\bibinfo{author}{\bibfnamefont{B.}~\bibnamefont{Hao}}, \bibinfo{author}{\bibfnamefont{T.}~\bibnamefont{Lattimore}}, \bibnamefont{and} \bibinfo{author}{\bibfnamefont{C.}~\bibnamefont{Szepesvari}}, \emph{\bibinfo{title}{Adaptive {{Exploration}} in {{Linear Contextual Bandit}}}} (\bibinfo{year}{2020}), \eprint{1910.06996}.

\bibitem[{\citenamefont{Rusmevichientong and Tsitsiklis}(2010)}]{rusmevichientongLinearlyParameterizedBandits2010}
\bibinfo{author}{\bibfnamefont{P.}~\bibnamefont{Rusmevichientong}} \bibnamefont{and} \bibinfo{author}{\bibfnamefont{J.~N.} \bibnamefont{Tsitsiklis}}, \bibinfo{journal}{Mathematics of OR} \textbf{\bibinfo{volume}{35}}, \bibinfo{pages}{395} (\bibinfo{year}{2010}).

\bibitem[{Note1()}]{Note1}
Note1, \bibinfo{note}{these results rely on an adjustment of the variance of the posterior distribution.}

\bibitem[{\citenamefont{Abeille and Lazaric}(2017)}]{abeilleLinearThompsonSampling2017}
\bibinfo{author}{\bibfnamefont{M.}~\bibnamefont{Abeille}} \bibnamefont{and} \bibinfo{author}{\bibfnamefont{A.}~\bibnamefont{Lazaric}}, \bibinfo{journal}{Electronic Journal of Statistics} \textbf{\bibinfo{volume}{11}}, \bibinfo{pages}{5165} (\bibinfo{year}{2017}).

\bibitem[{\citenamefont{Kirschner et~al.}(2021)\citenamefont{Kirschner, Lattimore, Vernade, and Szepesvari}}]{kirschnerAsymptoticallyOptimalInformationDirected2021}
\bibinfo{author}{\bibfnamefont{J.}~\bibnamefont{Kirschner}}, \bibinfo{author}{\bibfnamefont{T.}~\bibnamefont{Lattimore}}, \bibinfo{author}{\bibfnamefont{C.}~\bibnamefont{Vernade}}, \bibnamefont{and} \bibinfo{author}{\bibfnamefont{C.}~\bibnamefont{Szepesvari}}, in \emph{\bibinfo{booktitle}{Proceedings of {{Thirty Fourth Conference}} on {{Learning Theory}}}} (\bibinfo{publisher}{PMLR}, \bibinfo{year}{2021}), pp. \bibinfo{pages}{2777--2821}, ISSN \bibinfo{issn}{2640-3498}.

\bibitem[{\citenamefont{Kirschner and Krause}(2018)}]{kirschnerInformationDirectedSampling2018}
\bibinfo{author}{\bibfnamefont{J.}~\bibnamefont{Kirschner}} \bibnamefont{and} \bibinfo{author}{\bibfnamefont{A.}~\bibnamefont{Krause}}, in \emph{\bibinfo{booktitle}{Proceedings of the 31st {{Conference On Learning Theory}}}} (\bibinfo{publisher}{PMLR}, \bibinfo{year}{2018}), pp. \bibinfo{pages}{358--384}, ISSN \bibinfo{issn}{2640-3498}.

\bibitem[{\citenamefont{Kirschner et~al.}(2020)\citenamefont{Kirschner, Lattimore, and Krause}}]{kirschnerInformationDirectedSampling2020}
\bibinfo{author}{\bibfnamefont{J.}~\bibnamefont{Kirschner}}, \bibinfo{author}{\bibfnamefont{T.}~\bibnamefont{Lattimore}}, \bibnamefont{and} \bibinfo{author}{\bibfnamefont{A.}~\bibnamefont{Krause}}, in \emph{\bibinfo{booktitle}{Proceedings of {{Thirty Third Conference}} on {{Learning Theory}}}} (\bibinfo{publisher}{PMLR}, \bibinfo{year}{2020}), pp. \bibinfo{pages}{2328--2369}, ISSN \bibinfo{issn}{2640-3498}.

\bibitem[{\citenamefont{Bayati et~al.}(2020)\citenamefont{Bayati, Hamidi, Johari, and Khosravi}}]{bayatiUnreasonableEffectivenessGreedy2020}
\bibinfo{author}{\bibfnamefont{M.}~\bibnamefont{Bayati}}, \bibinfo{author}{\bibfnamefont{N.}~\bibnamefont{Hamidi}}, \bibinfo{author}{\bibfnamefont{R.}~\bibnamefont{Johari}}, \bibnamefont{and} \bibinfo{author}{\bibfnamefont{K.}~\bibnamefont{Khosravi}}, in \emph{\bibinfo{booktitle}{Advances in {{Neural Information Processing Systems}}}}, edited by \bibinfo{editor}{\bibfnamefont{H.}~\bibnamefont{Larochelle}}, \bibinfo{editor}{\bibfnamefont{M.}~\bibnamefont{Ranzato}}, \bibinfo{editor}{\bibfnamefont{R.}~\bibnamefont{Hadsell}}, \bibinfo{editor}{\bibfnamefont{M.~F.} \bibnamefont{Balcan}}, \bibnamefont{and} \bibinfo{editor}{\bibfnamefont{H.}~\bibnamefont{Lin}} (\bibinfo{publisher}{Curran Associates, Inc.}, \bibinfo{year}{2020}), vol.~\bibinfo{volume}{33}, pp. \bibinfo{pages}{1713--1723}.

\bibitem[{\citenamefont{Jamieson and Jain}(2018)}]{jamiesonBanditApproachMultiple2018}
\bibinfo{author}{\bibfnamefont{K.}~\bibnamefont{Jamieson}} \bibnamefont{and} \bibinfo{author}{\bibfnamefont{L.}~\bibnamefont{Jain}}, in \emph{\bibinfo{booktitle}{Proceedings of the 32nd {{International Conference}} on {{Neural Information Processing Systems}}}} (\bibinfo{publisher}{Curran Associates Inc.}, \bibinfo{address}{Red Hook, NY, USA}, \bibinfo{year}{2018}), {{NIPS}}'18, pp. \bibinfo{pages}{3664--3674}.

\bibitem[{Note2()}]{Note2}
Note2, \bibinfo{note}{the lower bounds holds for sufficiently high $T,K$ values and $1$-regular $\Gamma $ ($\mu _{\protect \mathrm {sup}}=1$), meaning that $\protect \mathcal {P}_\Gamma [ \mu > 1- \epsilon ] = O(\epsilon )$ as $\epsilon $ goes to $0$; see \cite {bayatiUnreasonableEffectivenessGreedy2020} for theorem details.}

\bibitem[{Note3()}]{Note3}
Note3, \bibinfo{note}{note that a sequential strategy has been proposed in \cite {bayatiUnreasonableEffectivenessGreedy2020}, but this strategy still depends on a fixed bound on the subset size.}

\bibitem[{\citenamefont{Andersen}(1954)}]{andersenFluctuationsSumsRandom1954}
\bibinfo{author}{\bibfnamefont{E.~S.} \bibnamefont{Andersen}}, \bibinfo{journal}{Mathematica Scandinavica} \textbf{\bibinfo{volume}{2}}, \bibinfo{pages}{195} (\bibinfo{year}{1954}), \eprint{24489033}.

\bibitem[{\citenamefont{Bauer et~al.}(1999)\citenamefont{Bauer, Godr{\`e}che, and Luck}}]{bauerStatisticsPersistentEvents1999}
\bibinfo{author}{\bibfnamefont{M.}~\bibnamefont{Bauer}}, \bibinfo{author}{\bibfnamefont{C.}~\bibnamefont{Godr{\`e}che}}, \bibnamefont{and} \bibinfo{author}{\bibfnamefont{J.~M.} \bibnamefont{Luck}}, \bibinfo{journal}{Journal of Statistical Physics} \textbf{\bibinfo{volume}{96}}, \bibinfo{pages}{963} (\bibinfo{year}{1999}).

\bibitem[{\citenamefont{Le~Doussal and Wiese}(2009)}]{ledoussalDrivenParticleRandom2009}
\bibinfo{author}{\bibfnamefont{P.}~\bibnamefont{Le~Doussal}} \bibnamefont{and} \bibinfo{author}{\bibfnamefont{K.~J.} \bibnamefont{Wiese}}, \bibinfo{journal}{Phys. Rev. E} \textbf{\bibinfo{volume}{79}}, \bibinfo{pages}{051105} (\bibinfo{year}{2009}).

\end{thebibliography}

\begin{titlepage}
  \vspace{1cm}
  \centering{\Large{{\bf Supplemental material}}}
  \clearpage
\end{titlepage}

\setcounter{figure}{0}
\setcounter{equation}{0}
\renewcommand{\figurename}{{SUPPLEMENTARY FIG.}}
\renewcommand{\thefigure}{{S\arabic{figure}}}
\renewcommand{\thesection}{{S\arabic{section}}}
\renewcommand{\theequation}{{S\arabic{equation}}}
\renewcommand{\tablename}{{SUPPLEMENTARY TABLE}}
\renewcommand{\thetable}{{S\arabic{table}}}

\begin{widetext}

\section{Details of Explored-M design}\label{Supp:SecExplore}

We provide all derivation details leading to $\algonameexp$ expression given in the main text.

\subsection{Link between information maximization and probability maximization}

Here, we briefly comment on the decision to switch from entropy increments to probability increments in the arm selection process of our $\algonameexp$ algorithm. The associated entropy for all arms, effectively well-separated by $\meanb$, is given by:
\begin{equation}
    S_{\mathrm{top}} = -\Ptopset  \ln  \Ptopset - (1- \Ptopset) \ln( 1-  \Ptopset),
\end{equation}
with  $\Ptopset = \prod_{i \in \topset} \Prob \left( \mu_i \geq \meanb - \frac{\epsilon}{2} \right) $. The information gain relative to arm $i \in \topset $ thus reads.
\begin{equation}\label{supp:topkentropy1}
   \Delta S_{\mathrm{top}} = -\Delta \Ptopset \left[ \ln(\Ptopset) - \ln(1-\Ptopset) \right].
\end{equation}
For $\Ptopset > 1/2$, the sign of $\Delta S_{\mathrm{top}}$ is given by $\Delta \Ptopset$, meaning that minimizing the entropy resumes to selecting the arm maximizing $\Delta \Ptopset$. At short times, one may face events with $\Delta \Ptopset < 1/2$. For such cases, direct entropy minimization will be inaccurate since \cref{supp:topkentropy1} ignores that arms may switch from $\topset$ to $\botset$, resulting in additional variations in $\Ptopset$ not caught in \eqref{supp:topkentropy1}. We thus assume to minimize to select arms  maximizing $\Delta \Ptopset | \ln(\Ptopset) - \ln(1-\Ptopset) |$ increment. The same argument can be drawn for $\Pbotset$.
\subsection{Details on increment derivation}\label{suppSecPexp}

We start by reminding $\Ptopset$ expression 

\begin{equation}\label{suppeq:explorePtop}
\Ptopset = \prod_{i \in \topset} \Prob \left( \mu_i \geq \meanb - \frac{\epsilon}{2} | \Ht \right),
\end{equation}

By leveraging the arms' independence, the product enables a straightforward computation of the increments along each arm 
\begin{equation}
\Delta_{i \in \topset } \Ptopset  = \frac{ \Delta \Prob \left( \mu_i \geq \meanb - \frac{\epsilon}{2} | \Ht \right)}{\Prob \left( \mu_i \geq \meanb - \frac{\epsilon}{2} | \Ht \right)} \Ptopset.
\end{equation}
Since all gradients can be factorized by $\Pbotset \Ptopset$, the decision criteria reduce to a comparison of the relative gradients of the cumulative distribution associated with each arm. Of note, a similar approach can be derived using the discrete increments. We finally conclude by computing the gradient of the cumulative probability associated with the arm $i$. It decomposes into partial derivatives along the empirical mean $\meanh{i}$ and the number of draws $\loct{i}$. We first notice that the partial derivative along $\meanh{i}$ cancels by symmetry. Then we consider the derivative along the number of draws.

For a Gaussian reward with unit variance, we have 

\begin{equation}
\begin{split}
\Prob \left( \mu_i \geq \meanb - \frac{\epsilon}{2} | \Ht \right) &= 1 - \frac{1}{2} \left[ 1 + \erf \left( \sqrt{\frac{\loct{i}}{2}} \left[ \meanb - \frac{\epsilon}{2} - \meanh{i}\right] \right) \right] \\
&= \frac{1}{2} \left[ 1 + \erf \left( \sqrt{\frac{\loct{i}}{2}} \left[ \meanh{i} - \meanb + \frac{\epsilon}{2} \right] \right) \right] 
\end{split}
\end{equation}
and 
\begin{equation}\label{suppeq:Pbotexp}
\begin{split}
 \Delta \Prob \left( \mu_i \geq \meanb - \frac{\epsilon}{2} | \Ht \right) &= \frac{\partial }{\partial \loct{i}  } \Prob \left( \mu_i \geq \meanb - \frac{\epsilon}{2} | \Ht \right) \\
 & =\frac{\meanh{i} - \meanb + \frac{\epsilon}{2} }{2 \sqrt{2 \pi \loct{i}}  }\exp \left( -\loct{i} \frac{ (\meanb - \frac{\epsilon}{2}  - \meanh{i})^2}{2} \right)
\end{split}
\end{equation}
Replicating this reasoning for $\Pbotset$ leads to
\begin{equation}\label{suppeq:Ptopexp}
\begin{split}
 \Delta \Prob \left( \mu_i \leq \meanb + \frac{\epsilon}{2} | \Ht \right) &= \frac{\partial }{\partial \loct{i}  } \Prob \left( \mu_i \geq \meanb - \frac{\epsilon}{2} | \Ht \right) \\
 &= \frac{\partial }{\partial \loct{i}  } \left( \frac{1}{2} \left[ 1 + \erf \left( \sqrt{\frac{\loct{i}}{2}} \left[   \meanb + \frac{\epsilon}{2} - \meanh{i} \right] \right) \right] \right)\\
 & =\frac{\meanb + \frac{\epsilon}{2} - \meanh{i} }{2 \sqrt{2 \pi \loct{i}}  }\exp \left( -\loct{i} \frac{ (\meanb + \frac{\epsilon}{2} - \meanh{i})^2}{2} \right)
\end{split}
\end{equation}

\subsection{Details on $\meanb$ derivation}

Here, we provide additional details on the derivation steps leading to $\meanb$ expression in the main text.  Our goal is to derive a tractable expression of an appropriate cutoff between the subset of arms with the current $\tk-$ highest empirical means, $\topset$, with a minimal empirical min $\topmin = \max (\botsetmeanh)$ and the remaining arms of maximal mean $ \topset$, $\botmax = \min(\botsetmeanh)$.

An optimal $\meanb$ value should permit the algorithms to successfully stop in a minimal amount of samples denoted $\tau$ in the following. For a given set of observed empirical rewards $\topsetmean$, $\botsetmean$ and neglecting their future variations, an oracle strategy will verify 
\begin{equation}\label{exploremoracle}
\meanbo = \underset{\theta \in [\botmax, \topmin]}{\argmin} \left( \tau = \sum_{i} \loctau{i}  \ \text{s.t.} \  \prod_{i \in \botset} \Prob \left( \mu_i \leq \theta + \frac{\epsilon}{2} | \meanh{i}, \loctau{i} \right) \prod_{i \in \topset} \Prob \left( \mu_i \geq \theta - \frac{\epsilon}{2}| \meanh{i}, \loctau{i}  \right) > 1-\delta \right).
\end{equation}
Here, $\meanbo$ is the optimal value that minimizes $\tau$ while satisfying the stopping criteria, and $\{\loctau{1} .. \loct{n}\}$ are the local times of each arm at the stopping time. Probabilities in \cref{exploremoracle} depend on the hypothetical number of samples for each arm, $\loctau{i}$ which differs by the current individual arm time $\loct{i}$ for $\Ptopset$ and $\Pbotset$. At short times, $\loctau{i} \gg \loct{i}$ because it takes a certain number of samples before meeting the conditions in \cref{exploremoracle}. However, by selecting a $\meanb$ close to $\meanbo$, the algorithm should adjust its strategy to match $\loct{i}$ with $\loctau{i}$ as closely as possible.

Unfortunately, the expression for $\meanbo$ becomes intractable with more than two arms. To derive a simplified expression, we approximate $\topset$ as a fictive set of $\tk$ arms, all with the same empirical mean, $\topmin$, and the same number of pulls, $t_{\mathrm{top}}$. Because all these fictive arms share the same empirical mean and pull count, the posterior distribution of their minimal mean value can be approximated  by a Gumbel law \cite{majumdarExtremeValueStatistics2020a} parametrized by $(\topmin, t_{\mathrm{top}}, \tk)$. 
\begin{equation}\label{sup:exploremgumbeltop}
\Prob \left( \underset{i \in \llbracket1.. \tk \rrbracket }{\min}( \mu_i  | \meanh{i}=\topmin, \loct{i} = t_{\mathrm{top}}) \geq \theta  \right) = \exp \left[ - \exp \left(  \sqrt{2\ln(\tk)} \left[  \sqrt{ t_{\mathrm{top}}} (\theta - \topmin) + \sqrt{2\ln(\tk)} \right] \right) \right].
\end{equation}
This approximation holds for large set values $\tk$, which are the most complicated cases for which relying on an efficient $\meanb$ value matters the most. By replicating the previous scheme for $\botset$, we obtain a Gumbel law parameterized by $(\botmax, t_{\mathrm{bot}}, \retk=\totk -\tk)$ with a cumulative distribution reading :
\begin{equation}\label{sup:exploremgumbelbot}
\Prob \left( \underset{i \in \llbracket1.. \retk \rrbracket }{\max}( \mu_i  | \meanh{i}=\botmax, \loct{i} = t_{\mathrm{bot}}) \leq \theta  \right) = \exp \left[ - \exp \left(  - \sqrt{2\ln(\retk)} \left[  \sqrt{ t_{\mathrm{bot}}} (\theta - \botmax) - \sqrt{2\ln(\retk)} \right] \right) \right].
\end{equation}
Let us note that the centering parameters $\sqrt{2\ln(\tk)}$ and  $\sqrt{2\ln(\retk)}$ result from the Gaussian property of the reward distribution.  Hence, we define $\meanb$ as simultaneously verifying
\begin{equation}\label{sup:exploremcondtop}
\Prob \left( \underset{i \in \llbracket1.. \tk \rrbracket }{\min}( \mu_i  | \meanh{i}=\topmin, \loct{i} = t_{\mathrm{top}}) \geq \meanb  -\frac{\epsilon}{2}  \right) =  1 - \frac{\delta}{2},
\end{equation}
and 
\begin{equation}\label{sup:exploremcondbot}
\Prob \left( \underset{i \in \llbracket1.. \retk \rrbracket }{\max}( \mu_i  | \meanh{i}=\botmax, \loct{i} = t_{\mathrm{bot}}) \leq \meanb + \frac{\epsilon}{2} \right) =  1 - \frac{\delta}{2}.
\end{equation}
These events include all cases where the fictive subset is not $\epsilon$-optimal, upper-bounding the identification error probability by $\delta$. By injecting \cref{sup:exploremgumbeltop,sup:exploremgumbelbot} expressions within \cref{sup:exploremcondtop,sup:exploremgumbelbot} one obtains:
%
%
\begin{equation}\label{sup:explorem_meantop}
 \meanb  =  \frac{\epsilon}{2} + \topmin - \frac{1}{  \sqrt{ t_{\mathrm{top}}}} \left[ \sqrt{2\ln(\tk)} - \frac{ \ln( - \ln(1 - \frac{\delta}{2}))}{ \sqrt{2\ln(\tk)}} \right],
\end{equation}
and 
\begin{equation}\label{sup:explorem_meanbot}
 \meanb  =  -\frac{\epsilon}{2} + \botmax + \frac{1}{  \sqrt{ t_{\mathrm{bot}}}} \left[ \sqrt{2\ln(\retk)} -  \frac{ \ln( - \ln(1 - \frac{\delta}{2}))}{ \sqrt{2\ln(\retk)}} \right].
\end{equation}
%
%
%
By combining \cref{sup:explorem_meantop,sup:exploremcondbot} one obtains 
\begin{equation}\label{sup:explorem_condf}
  \epsilon+ \topmin  - \botmax =  \frac{1}{  \sqrt{ t_{\mathrm{top}}}} \left[ \sqrt{2\ln(\tk)} - \frac{ \ln( - \ln(1 - \frac{\delta}{2}))}{ \sqrt{2\ln(\tk)}} \right]  + \frac{1}{  \sqrt{ t_{\mathrm{bot}}}} \left[ \sqrt{2\ln(\retk)} -  \frac{ \ln( - \ln(1 - \frac{\delta}{2}))}{ \sqrt{2\ln(\retk)}} \right].
\end{equation}
We finally look for $t_{\mathrm{top}}$ and $t_{\mathrm{bot}}$ verifying \cref{sup:explorem_condf} but leading to a minimal amount of time $ \tk t_{\mathrm{top}}$ + $\retk t_{\mathrm{bot}}$. By labeling
\begin{equation}\label{supp:coefftopep}
\coefftop = \left[ \sqrt{2\ln(\tk)} - \frac{ \ln( - \ln(1 - \frac{\delta}{2}))}{ \sqrt{2\ln(\tk)}} \right], \: \; \coeffbot = \left[ \sqrt{2\ln(\retk)} - \frac{ \ln( - \ln(1 - \frac{\delta}{2}))}{ \sqrt{2\ln(\retk)}} \right], \: \: M = \epsilon+ \topmin  - \botmax
\end{equation}
we obtain 
\begin{equation}
t_{\mathrm{top}} =  \left[ \frac{\coefftop}{ M - \frac{\coeffbot}{\sqrt{t_{\mathrm{bot}}}}} \right]^2,
\end{equation}
and  
\begin{equation}
\begin{split}
t_{\mathrm{bot}} &= \frac{\coeffbot}{M^2} + \left( \frac{\coeffbot^2 \coefftop^4 \tk^2}{M^6 \retk^2}\right)^\frac{1}{3} + \frac{2 M^2 \retk}{\coefftop^2 \tk} \left( \frac{\coeffbot^2 \coefftop^4 \tk^2}{M^6 \retk^2}\right)^\frac{2}{3},
\end{split}
\end{equation}
which simplifies into
\begin{equation}\label{supp:topktopexpsim}
\begin{split}
t_{\mathrm{bot}} &= \frac{\coeffbot}{M^2} + \left( \frac{\coeffbot^2 \coefftop^4 \tk^2}{M^6 \retk^2}\right)^\frac{1}{3}  \left( 1 + \frac{2 d^\frac{2}{3} \coeffbot^\frac{2}{3}}{\tk^\frac{1}{3} \coefftop^\frac{2}{3}} \right).
\end{split}
\end{equation}
Inserting \cref{supp:topktopexpsim} in \cref{supp:coefftopep} gives the desired analytical expression of $\meanb$
\begin{equation}\label{sup:explorem_lastexprmeanb}
 \meanb  =  -\frac{\epsilon}{2} + \botmax + \coeffbot \left[ \frac{\coeffbot}{M^2} + \left( \frac{\coeffbot^2 \coefftop^4 \tk^2}{M^6 \retk^2}\right)^\frac{1}{3} + \frac{2 M^2 \retk}{\coefftop^2 \tk} \left( \frac{\coeffbot^2 \coefftop^4 \tk^2}{M^6 \retk^2}\right)^\frac{2}{3}
 \right]^{-\frac{1}{2}}.
\end{equation}
Note that our approximation scheme is mainly independent of the reward distribution, as it only involves in $\coeffbot$ and $\coefftop$ expressions. To extend our results to various reward distributions, one simply needs to adjust the Gumbel law by providing the corresponding centering parameters (here $\sqrt{2\ln(\tk)}$ for Gaussian rewards). Leveraging on scaling proposed in \cite{majumdarExtremeValueStatistics2020a} for any distribution exhibiting a large $x$ tail $p(x) \sim \exp( -x^\gamma) $, the adjusted Gumbel law distributions read
\begin{equation}\label{sup:exploremgumbeltop_general}
\Prob \left( \underset{i \in \llbracket1.. \tk \rrbracket }{\min}( \mu_i  | \meanh{i}=\topmin, \loct{i} = t_{\mathrm{top}}) \geq \theta  \right) = \exp \left[ - \exp \left( \gamma \ln(\tk)^{1 - \frac{1}{\gamma}} \left[  \sqrt{ t_{\mathrm{top}}} (\theta - \topmin) + \ln(\tk)^{\frac{1}{\gamma}} \right] \right) \right]
\end{equation}
and
\begin{equation}\label{sup:exploremgumbelbot_general}
\Prob \left( \underset{i \in \llbracket1.. \retk \rrbracket }{\max}( \mu_i  | \meanh{i}=\botmax, \loct{i} = t_{\mathrm{bot}}) \leq \theta  \right) = \exp \left[ - \exp \left( \gamma \ln(\retk)^{1 - \frac{1}{\gamma}} \left[  \sqrt{ t_{\mathrm{bot}}} (\theta - \botmax) + \ln(\retk)^{\frac{1}{\gamma}} \right] \right) \right],
\end{equation}
only affecting $\coeffbot$ and $\coefftop$ expressions. Still, keeping the Gaussian approximation within $\meanb$ should remain reasonable for non-heavy-tailed reward distributions.

\section{Details of \algonamelin{ } for linear bandits}\label{Supp:SecLinear}

Here we provide derivations and simplifications details regarding our algorithm for the linear bandit setting.

\subsection{Note on posterior distribution associated to $\Astar$ value}

We first seek to derive the posterior probability distribution of $\Astar$ value given the history $\Ht$, which consists of past pairs of chosen arms and their corresponding rewards ($\Ain$, $\rewardn$ for $n \in [\![0, t]\!]$). We also rely on a Gaussian prior for $\Astar$ value with a $L^2$ norm:
\begin{equation}\label{supp:eqpriorlin}
p(\Astar = \theta) = \mathcal{N}^{-1} \exp \left( - \frac{\lambda}{2} \sum^d_{i} \theta^{2}_i \right),
\end{equation}
where $d$ is the arm space dimension. Choosing $\lambda>0$ serves two purposes: first, it ensures that the covariance matrix of the associated posterior distribution is invertible even when no arms have been drawn. Second, it focuses the inference on more realistic optimal values by expecting a mean reward close to the unit norm. According to Bayes' theorem, the posterior distribution reads
\begin{equation}\begin{split}\label{supp:eqlinpost}
p(\Astar = \theta | \Ht)  &= p(\Ht | \Astar = \theta ) p(\Astar = \theta) p(\Ht)^{-1} \\
&=  p(\Ht)^{-1} \frac{\mathcal{N}^{-1} e^{  - \frac{\lambda}{2} \sum^d_{i} \theta^{2}_i } }{(2\pi)^{t/2}}  \exp \left(  - \frac{1}{2} \sum^t_{n=0} \left[ \theta^{\tr} \Ain  - \rewardn \right]^2 \right) \\
&=p(\Ht)^{-1} \frac{\mathcal{N}^{-1} e^{ - \frac{1 }{2} \sum^t_{n=0} \rewardn^2 }  }{(2\pi)^{t/2}} \exp \left( \theta^{\tr} \sum^t_{n=0} \rewardn \Ain  - \frac{1}{2} \theta^{\tr} \left[ \sum^t_{n=0} \Ain \Ain^{\tr} + \lambda I_d \right] \theta \right) \\
&= \frac{ \exp \left( - \frac{1}{2}(\theta- \thetaemp )^{\tr} \bconfmt (\theta- \thetaemp)  \right)} {\sqrt{(2 \pi)^d \det \bconfmt^{-1} }}
\end{split}
\end{equation}
where we defined $ \bconfmt = \lambda I_d +  \sum_\horin^{t} \Ain \Ain^{\tr}$ and $\thetaemp = \bconfmt^{-1}\sum_{\horin}^{t} \rewardn  \Ain$. The final step relies on the distribution being normalized. \cref{supp:eqlinpost} holds for Gaussian rewards and one recognizes a multivariate Gaussian distribution. \cref{supp:eqlinpost} forms the basis for the entropy calculations that underpin our algorithm.
\subsection{Details of entropy increments for linear bandits \algonamelin{ }}

Here we outline the key steps for computing the entropy increments discussed in the main text.

\subsubsection{Entropy assigned to the best available empirical arm}

We first consider the best available empirical arm at time $t$ defined as $ \Amaxa = \underset{a \in \decset}{\max}  \langle \Astar, a \rangle$. We assign to this arm the information gain of $\Astar$ value which posterior is given by \cref{supp:eqlinpost}. Because \cref{supp:eqlinpost} distribution is a Gaussian multivariate, the associated entropy reads
\begin{equation}\label{supp:eqSastar}
\begin{split}
S &= - \int_{\mathbb{R}^d} p(\Astar = \theta | \Ht) \ln p(\Astar = \theta | \Ht) \df \theta \\
   &= \frac{1}{2} \ln(\det \bconfmt^{-1})  + \frac{d}{2} [ 1 + \ln(2\pi)] 
\end{split}
\end{equation}
The resulting increment, hence reads:
\begin{equation}
\begin{split}\label{supp:DeltaSastar}
\Delta  S  &=- \frac{1}{2} \ln \left[ \frac{\det \left( \bconfmt + \Amaxa \Amaxa^{\tr} \right)}{ \det \left( \bconfmt  \right)}  \right] \\
 &= -\frac{1}{2} \ln \left[ \frac{ \left( 1 + \Amaxa^{\tr} \bconfmt^{-1} \Amaxa \right)\det \left(\bconfmt \right) }{ \det \left(\bconfmt \right)}  \right] \\
  &= -\frac{1}{2} \ln  \left( 1 + \Amaxa^{\tr} \bconfmt^{-1} \Amaxa \right) \\
\end{split}
\end{equation}
where we used the matrix determinant lemma with $\bconfmt^{-1}$ and $\Amaxa \Amaxa^{\tr}$. 

\subsubsection{Entropy assigned to the empirical suboptimal arms}

Next, we consider the entropy assigned to the remaining suboptimal actions of $\decset$. For each arm, denoted $\Ait$, we assign the information gain regarding the event $\Astar^{\tr} (\Amaxa - \Ait) > 0$ of probability $C_{\Amaxa, \Ait}$ :
\begin{equation}\label{supp:eqCAmaxa}
C_{\Amaxa > \Ait} =  \frac{1}{2} \left[  1 + \erf \left( \frac{\thetaemp^{\tr} [\Amaxa - \Ait]}{ \sqrt{2 [\Amaxa - \Ait]^{\tr} \bconfmt^{-1}  [\Amaxa - \Ait]^{\tr}}}\right)\right],
\end{equation}
with its associated entropy 
\begin{equation}
S_{C, \Ait} = - C_{\Amaxa > \Ait} \ln C_{\Amaxa > \Ait}  - (1-C_{\Amaxa > \Ait} ) \ln \left( 1 - C_{\Amaxa > \Ait} \right).
\end{equation}
To get a simplified expression we made a first-order expansion of $C_{\Amaxa > \Ait}$ and of its entropy. As for AIM approaches, we only focus on the asymptotic regime where the probability of a seemingly suboptimal arm being optimal is a rare event, in other words when  $C_{\Amaxa > \Ait} \approx 1 $. Surprisingly, this approximation is still accurate enough outside this asymptotic regime to provide a high-performance decision scheme.
We first make a first-order expansion in the regime $C_{\Amaxa > \Ait} \approx 1$. Denoting the inner term of the error function in $\cref{supp:eqCAmaxa}$ as $\frac{\Meanerf}{ \sqrt{2\Sigerf}}$ with $\Meanerf = \thetaemp^{\tr} [\Amaxa - \Ait]$ and $\Sigerf = [\Amaxa - \Ait]^{\tr} \bconfmt^{-1}  [\Amaxa - \Ait]^{\tr}$ leads to

%
\begin{equation}
C_{\Amaxa > \Ait} =  \frac{e^{-\frac{\Meanerf^2}{2 \Sigerf}}}{2 \frac{\Meanerf}{2 \sqrt{\Sigerf}} \sqrt{\pi}} \left(1 + o(1) \right)
\end{equation}
and 
%
\begin{equation}\label{supp:eqsaitapprox}
\tilde{S}_{C,\Ait} \approx  \frac{\Meanerf}{ \sqrt{\Sigerf}} \frac{e^{-\frac{\Meanerf^2}{ 2\Sigerf}}}{2 \sqrt{\pi}} 
\end{equation}
Applying \cref{supp:DeltaSastar} to $S$ for $\Ait$ and multiplying it with \cref{supp:eqsaitapprox} viewed as a weight leads to the final expression given in the main text.

\section{Details of \algonamesemany{ for the many-armed settings} }\label{Supp:SecMany}

Here, we recapitulate all the steps leading to the analytical expression constitutive of our algorithm in the many-arms setting. We stress that it involves considerably simplifying our functional expression to get the final form of \algonamesemany{ }. We will also evoke extensions to more general prior distributions.

We start by commenting on the partition scheme and their related approximations leading to both upcoming regret expressions given in the main text. The regret induced by solely selecting the currently better empirical arm is given by : 
\begin{equation}\label{supp:eqmanyarms1}
\begin{split}
\Rgr &= \tau\left( \mustar - \int_{\mubmin}^{\mubmax} \mu \, p( \mu_{\maxa} = \mu | \mumaxt, \loct{\maxa}, \Gamma) \df \mu \right)\\
&\approx \tau ( \mustar -  \mumaxt ),
\end{split}
\end{equation}
where $\retime$ denotes the remaining time and the integral spans over the mean reward of the current best empirical mean based on its history.
To be more accurate, the resulting regret of a policy reevaluating the current best empirical arm at each time step should also account for the possibility that the best empirical arm may change over time. Yet, approximating this policy by the simpler strategy given by \cref{supp:eqmanyarms1} significantly simplifies the derivation without introducing undesirable behavior.  We also approximate the mean posterior value by the empirical mean in \cref{supp:eqmanyarms1} which provides a greedy strategy to explore the most promising arm. Next, we derive the second upcoming regret expression given in the main text. We make a partition based on how the mean reward of the newly explored arm behaves compared to $\mumaxt$ :
\begin{equation}\label{supp:eqmanyarms2}
\begin{split}
\Rexp &=  \tau \int_{\mumaxt}^{\mubmax}  \muprior(\mu)(\mustar - \mu) \df \mu  \: \: + \\ &\int_{\mubmin}^{\mumaxt} \muprior(\mu) \left[  (\mustar -  \mu)\tfpt  + (\mustar - \mumaxt) \left( \tau - \tfpt \right)  \right] \df \mu.
\end{split}
\end{equation}
The first term approximates cases where the new arm is better than $\mumaxt$. Here, we neglect cases where the new arm's mean falls below $\mumaxt$. The second term accounts for cases where the new arm mean is lower than $\mumaxt$. Our policy will continue pulling such unfavorable arms until their mean reward falls below $\mumaxt$, an event occurring on average after $\tfpt$ time steps. Finally, for consistency with \cref{supp:eqmanyarms1}, we approximate the regret induced by solely selecting the current best empirical arm using $\mumaxt$. An efficient policy should avoid allocating exploration time to previously explored arms that already appear to be suboptimal, even if there is a finite probability that they may be the better choice. Given the large number of arms, our policy will either select arms with empirical means greater than or equal to $\mumaxt$ or explore an untested arm. Indeed, if the current best empirical mean starts to decline, there will always be a nearby arm with a similar empirical mean. Iterating this process, it becomes clear that the algorithm will end up finding an arm with a mean close to, if not better than, $\mumaxt$. This observation justifies that our greedy policy can be effectively approximated by $\mumaxt$. Therefore, we assume that cases where a better arm falls below $\mumaxt$ can be neglected, as such an arm will no longer impact the regret. Such probability may also be derived but doesn't seem to improve the algorithm's efficiency. \\

We now present the last steps to get  a tractable expression, i.e., obtaining $\tfpt$ scaling. We label by $i$ the new tested arm :
\begin{equation}
    \tfpt = \min( \E{ \min \left( s > t \mid \mumaxt > \meanh{i}(s) \right) | \mu_{i} = \mu},\tau), 
\end{equation}
where arm $i$ is drawn at each time until either $\meanh{i}(s)$ falls below $\mumaxt$ or the time horizon is reached. Since the successive rewards are independent from each other, $\tfpt$ can be modeled by the first time where $\sum_{i}^n X_i < n \mumaxt  $ with $X_i$ being independent identically distributed according to the reward distribution parameterized by $\mu$. It thus resumes as the mean first-passage time of a biased random walk dropping below $\mumaxt$. For an asymmetric jump distribution $\tfpt$ may be deduced from a generalized version of the Sparre-Andersen theorem \cite{andersenFluctuationsSumsRandom1954, bauerStatisticsPersistentEvents1999, ledoussalDrivenParticleRandom2009}. By defining $q_+(n= \mathrm{Prob}[ x_n \geq 0, x_{n-1} \geq0,...|x_0=0] $, and its associated generative series $\tilde{q}_+(s) = \sum_{s} q_+(n) s^n$. It states that 
\begin{equation}\label{supp:manyeqspare}
\tilde{q}_+(s) = \exp \left[ \sum_{1}^\infty \frac{s^n}{n} \mathrm{Prob}\left( \sum_{i}^n X_i \geq \mumaxt n \right) \right].
\end{equation}
By noticing 
\begin{equation}
\begin{split}
\tfpt &= \sum_{n=1}^\infty n [q_+(n-1)-q_+(n)] \\
&=  \sum_{n=1}^\infty q_+(n-1) + \sum_{n=1}^\infty (n-1) q_+(n-1) - \sum_{n=1}^\infty n q_+(n)\\
&= \tilde{q}_+(1)\\
\end{split}
\end{equation}
we obtain using \cref{supp:manyeqspare}
\begin{equation}\label{supp:manyarmeq_tpftgeneral}
\tfpt = \exp \left[ \sum_{1}^\infty \frac{1}{n} \mathrm{Prob}\left( \sum_{i}^n X_i \geq \mumaxt n \right) \right].
\end{equation}
This general expression holds for an asymmetric jump distribution, or equivalently, for a general reward distribution. To get a typical behavior, we approximate the jump distribution by a Gaussian of mean $\mu$ and unit variance. Applying it to \cref{supp:manyarmeq_tpftgeneral} gives :
\begin{equation}\label{supp:manyarmeq_2}
\tfpt = \exp \left[ \sum_{1}^\infty \frac{1}{2n} \mathrm{erfc}\left( \frac{\sqrt{n}}{\sqrt{2}} (\mumaxt-\mu) \right)   \right].
\end{equation}
There is no exact expression for \cref{supp:manyarmeq_2}, we hence focus on extracting the scaling of $\tfpt$ . We first focus on the limit case where  $\mu \sim \mumaxt$. In this regime, the erfc function will behave as $1/2$ until a cutoff $\sqrt{n_c}(\mumaxt-\mu)^{-1} \sim O(1)$. Hence, the leading order behaves as 
\begin{equation}\label{supp:manyarmeq_3}
\begin{split}
\tfpt &=  \exp \left[ \frac{1}{2} \ln \left( \frac{1}{(\mumaxt-\mu)^2} \right) + O(1) \right]\\
& =O\left( [\mumaxt-\mu]^{-1} \right).\\
\end{split}
\end{equation}
This regime includes all cases where the probability of observing diverging $\tfpt$ is the highest. In cases where $\mumaxt-\mu = O(1)$, the algorithm will mostly come back to $\mumaxt$ after a few steps, which is also given by \cref{supp:manyarmeq_3}. Hence, we assume to extend this approximation in all regimes, leading to the $\tfpt$ scaling given in the main text.
\begin{equation}\label{supp:manyarmeq_4}
\tfpt(\mu) = \frac{\almprefactor}{\mumaxt-\mu}
\end{equation}
%
Combining \cref{supp:eqmanyarms1,supp:eqmanyarms1,supp:manyarmeq_4} altogether gives the following decision criteria
\begin{equation}
\begin{split}
\Delta &= \Rexp - \Rgr  =  \tau \int_{\mumaxt}^{\mubmax}  \muprior(\mu)(\mustar - \mu) \df \mu  + \int_{\mubmin}^{\mumaxt} \muprior(\mu) \left[  (\mustar -  \mu)\tfpt  + (\mustar - \mumaxt) \left( \tau - \tfpt \right)  \right] \df \mu -
\tau (\mustar - \mumaxt)  \\
&=  \tau \int_{\mumaxt}^{\mubmax}  \muprior(\mu)(\mumaxt - \mu) \df \mu +  \int_{\mubmin}^{\mumaxt} \muprior(\mu)  (\mumaxt -  \mu)\tfpt \df \mu\\
&=  \tau \int_{\mumaxt}^{\mubmax}  \muprior(\mu)(\mumaxt - \mu) \df \mu + \almprefactor  \int_{\mubmin}^{\mumaxt} \muprior(\mu) \df \mu.
\end{split}
\end{equation}
For a uniform prior, it resumes as the expression given in the main text :
\begin{equation}
\begin{split}
\Delta &=   \almprefactor \frac{\mumaxt - \mubmin}{\mubmax - \mubmin} - \frac{\tau}{2} \frac{(\mubmax - \mumaxt)^2}{ \mubmax-\mubmin}.
\end{split}
\end{equation}
Finally for a uniform prior on $[0,1]$, it simplifies into :
\begin{equation}\label{supp:manyeqfinaldelta}
\Delta  =  \almprefactor \mumaxt - \tau \frac{(1-\mumaxt)^2}{2} 
\end{equation}
\subsubsection{Heuristic arguments supporting asymptotic optimality}

To further support the empirical evidence of \algonamesemany{} optimality, we consider the asymptotic regime where the algorithms have just converged on a promising arm, given by $\Delta \approx 0$ in \cref{supp:manyeqfinaldelta} which is fulfilled when :
\begin{equation}\label{supp:eqscalemany}
    \retime (1- \mumaxt)^2 \sim O(1).
\end{equation}
We denote $t_f = T - \tau$ the time spent, and by $n_f$ the current number of tested arms. We assume that $n_f$ scales as $O(t_f)$ as long as the promising arm is not found. For a uniform prior, and assuming that the promising arm belongs to the best arms of the subset, the expected regret $(1- \mumaxt)$ will thus scale as $1-\mean{i} \sim 1/n_f \sim 1/t_f$. Injecting such scaling into \cref{supp:eqscalemany} leads to $\retime t_f^2 \sim 1$ reached for the first time when $t_f \sim \sqrt{\hori}$. Hence, the past regret will scale as  $t_f \sim \sqrt{\hori}$ and the resulting upcoming regret of exploiting $\mumaxt$ will scale as $\sqrt{\retime} \sim \sqrt{\hori}$ ensuring the expected optimal scaling. To derive a complete proof, it will be necessary to ensure that the algorithm doesn't fail to identify $\mumaxt$ when tested and not longer spend most of its time trying to identify $\mumaxt$ within the arm subset.


\section{Numerical experiments}\label{supp:Secnumericalexpe}

\subsection{Numerical settings}

\subsubsection{Explore-$m$ bandit setting}

We here consider Gaussian rewards with $\sigma=1$. The $\epsilon$-optimal subset size verifies $\tk= \totk/5$ with an increasing number of arms tested. In \cref{fig:1} \textbf{(a)} the mean rewards values are sampled from two disjoint ensembles $[0,0.8]$ and $[0.9,1]$ for the $\epsilon$-optimal subset, with confidence parameters $\epsilon=0.05$, $\delta=0.1$.  In \cref{fig:1} \textbf{(b)} suboptimal mean rewards values are all set to $0.7$ while $\epsilon$-optimal to $0.8$, with confidence parameters $\epsilon=0.05$, $\delta=0.02$. For both experiments, the time is averaged over $100$ runs with standard deviation indicated. Here, seeds are not shared through the algorithms within the same experiment.

\subsubsection{Linear bandit setting}

We here again consider rewards subject to Gaussian noise with $\sigma=1$. In the first experiment, the arms composing $\decset$ are sampled from a normal distribution in $\mathbb{R}^{10}$. $\Astar$ is assigned a unit norm. The seeds are taken independently for each algorithm. The regret, averaged over $100$ runs, already shows a clear and consistent standard deviation. In the second experiment, $\xi=0.1$ and the $\decset$ are constituted by two fixed sets  \(\{[1, 0, 0], [0, 1, 0], [1-\xi, 2\xi, 0]\}\), the second set \(\{[0, 0.6, 0.8], [0, 0, 1], [0, \xi/10, 1-\xi]\}\), presented to the agent with equal probability and $\xi=0.1$, finally \(\Astar = [1, 0, 1]\). The regret, averaged over $1000$ runs, already shows a clear and consistent standard deviation. Here, seeds are shared through the algorithms within the same experiment.

\subsubsection{Many armed bandit setting}

Here, we consider Bernoulli rewards with a uniform prior. In \textbf{(a)}, the parameters used are
 $\hori=10000$, $K=5000 \gg \sqrt{\hori}$ and $c=1$. The regret is averaged over $2000$ runs. In \textbf{(b)} the mean regret performance when the horizon is reached is given for distinct games with varying horizon length ($K=\hori \gg \sqrt{\hori}$ and $c=1$). The regret is averaged over $2000$ runs for each experiment. Here, seeds are shared through the algorithms within the same experiment.






\subsection{Overview of baselines algorithms}\label{supp:Secotheralgo}

We will here review the algorithm used for numerical comparison.

\subsubsection{Top-k identification: LUCB1}

We rely on the algorithm version first introduced in \cite{kalyanakrishnanPACSubsetSelection2012a}. It falls under the category of upper confidence bound (UCB) algorithms, which select the arm maximizing a proxy function typically defined as $F_i = \meanh{i} + C_i(\loct{i},t)$. In the specific case of subset selection, it reformulates as follow. We define $B_i$ as 

\begin{equation}
     C_i(\loct{i},t) = \sqrt{ \frac{1}{2 \loct{i}} \ln \left[ \frac{5 \totk t^4}{4 \delta} \right]}.
\end{equation}

After having sorted the arms according to their current empirical means and having formed $\topset$ and $\botset$, the algorithm selects two arms, denoted $l_{\mathrm{top}, t}$ and $h_{\mathrm{bot},t}$, respectively in $\topset$ and $\botset$ such that 

\begin{equation}
l_{\mathrm{top}, t} = \underset{i \in \topset}{\argmin} \left( \meanh{i} -  C_i(\loct{i},t)\right),
\end{equation}

and 

\begin{equation}
h_{\mathrm{bot},t} = \underset{i \in \botset}{\argmax} \left( \meanh{i} +  C_i(\loct{i},t)\right).
\end{equation}

Then, the algorithm stops if 

\begin{equation}
 \meanh{h_{\mathrm{bot},t}} +  C_{h_{\mathrm{bot},t}}(\loct{l_{\mathrm{bot}, t} },t) -\meanh{l_{\mathrm{top}, t} } +  C_{l_{\mathrm{top}, t}} (\loct{l_{\mathrm{top}, t} },t) < \epsilon,
\end{equation}

meaning that both sets are sufficiently differentiated. If the algorithm does not stop, both arms $l_{\mathrm{top}, t}$ and $h_{\mathrm{bot},t}$ are pulled, and the process is repeated.

\subsubsection{Linear bandit: LinUCB}

As for the subset selection, we provide algorithms falling into the  upper confidence bound (UCB) category. These algorithms select the arm maximizing a proxy function typically defined as $F_{i,t} = \meanh{i,t} + C_{i,t}$. In the linear case  $\meanh{i}$ stands for the empirical mean of drawing arm $i$ at $t$ and $C_i$ an associated confidence ellipsoid.
The first LinUCB-type algorithm considered in the numerical experiments, referred to as OFUL \cite{abbasi-yadkoriImprovedAlgorithmsLinear2011}, is given by

\begin{equation}
F_{i,t} =  \thetaemp^{\tr} \Ait + \left[ \sqrt{ d \ln \left( n (1+t) \right)} + 1 \right] \sqrt{ \Ait^{\tr} \bconfmt^{-1} \Ait  },
\end{equation}

where $d$ stands for the dimension of the arms space, $n$ the arm number at each time step, and $\bconfmt$ as $\thetaemp$ are defined as in the main text. The second version borrowed from \cite{lattimoreBanditAlgorithms2020}, denoted LinUCB here, reads

\begin{equation}
F_{i,t} =  \thetaemp^{\tr} \Ait + \left[ \sqrt{ 2 \ln(n) + \ln \left( |\bconfmt| \right) } + 1 \right] \sqrt{ \Ait^{\tr} \bconfmt^{-1} \Ait  },
\end{equation}

where $||$ denotes the determinant operator. We now briefly discuss several potential candidate algorithms that were not implemented in this work. One such algorithm is Lin-Thompson, a refined version of the popular Thompson sampling approach. However, the linear variant of Lin-Thompson, which comes with theoretical guarantees, has a critical drawback: its posterior distribution's variance is adjusted by a factor of $\ln(t)$ \cite{agrawalThompsonSamplingContextual2013}. This adjustment leads to over-exploration and consequently, significant regret losses, rendering the algorithm inadequate for meaningful performance comparisons. 
In addition to Lin-Thompson, we could consider comparing our algorithm with other state-of-the-art methods such as SOLID \cite{tirinzoniAsymptoticallyOptimalPrimaldual2020}, OSSB \cite{combesMinimalExplorationStructured2017a}, OAM \cite{haoAdaptiveExplorationLinear2020}, and IDS \cite{kirschnerAsymptoticallyOptimalInformationDirected2021}. Nevertheless, these algorithms suffer implementation complexities that are not tackled in our current work.

\subsubsection{Many-arms: SS-epsilon greedy}

The subsampled greedy algorithm for many-armed settings randomly selects a subset of $m$ arms and performs a greedy strategy after pulling all the arms once. Its pseudo code reads : 

\begin{itemize}
\item Select $m$ arms uniformly at random without replacement from $\kset$ and pull each arm once.
\item Until the end pulls $\armt = \argmaxK \meanh{i}$.
\end{itemize}

The size of the subset is the only parameter that is adjusted depending on the horizon sizes and the reward distribution. \cite{bayatiUnreasonableEffectivenessGreedy2020} proposes $m \sim T^{\beta/(1+\beta)}$ for $\beta$-regular prior, i.e., $\Prob_\muprior  [ \mu  > 1-\epsilon ] = O(\epsilon^\beta)$.

\subsection{Additional figures}

\begin{figure}[htpp]
\centering
\includegraphics[scale=0.7]{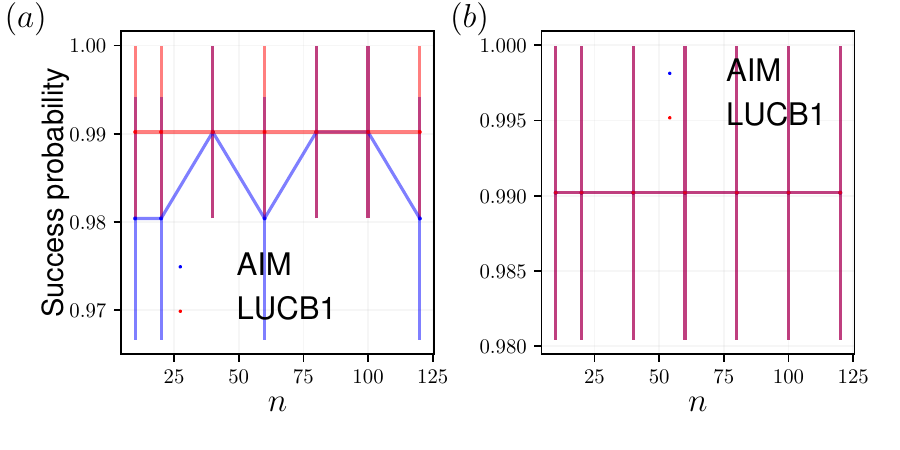}
\caption{Success probability of the Explore-$m$ experiments presented in the main text.
The probability of identifying the $\epsilon$-optimal subset is averaged over $100$ runs for different numbers of arms $\totk$. Results are shown for our algorithm (blue) and LUCB1 \cite{kalyanakrishnanPACSubsetSelection2012a} (red). The $\epsilon$-optimal subset size satisfies $\tk = \totk/5$.
(a) Mean reward values are sampled from two disjoint ensembles $[0, 0.7]$ and $[0.8, 1]$, with confidence parameters $\epsilon=0.05$ and $\delta=0.1$.
(b) Suboptimal mean reward values are fixed at $0.7$, while $\epsilon$-optimal rewards are set to $0.8$, with confidence parameters $\epsilon=0.05$ and $\delta=0.02$.
Both experiments demonstrate an empirical success probability exceeding the tolerance threshold $1-\delta$, validating the stopping time measured in the main text.
}
\label{figsupp:1pres}
\end{figure}

\begin{figure}[htpp]
\centering
\includegraphics[scale=0.8]{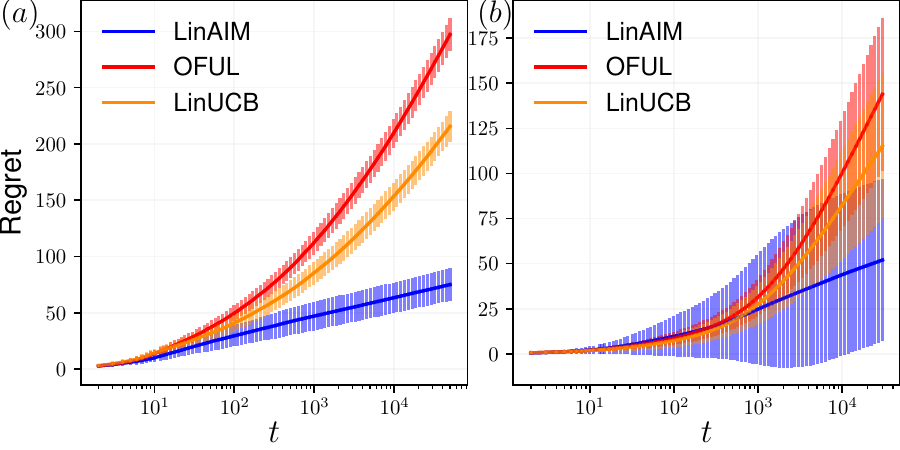}
\caption{ Mean regret for two linear bandit settings with Gaussian rewards. This figure is identical to the one given in the main text but with standard deviation shown with error bars. In blue our algorithm and in orange and red UCB algorithms adjusted to the linear settings. \textbf{(a)} Bandit setting with $10$ arms resampled at each time from a normal distribution in $\mathbb{R}^{10}$. \textbf{(b)} Toy problem borrowed from \cite{tirinzoniAsymptoticallyOptimalPrimaldual2020} with two contexts. Each context (detailed in the main text) is drawn with equal probability and $\xi=0.5$. Details of the algorithms, simulations and a focus on standard deviations are provided in \cref{supp:Secnumericalexpe}.}
\label{suppfig:2}
\end{figure}

\begin{figure}[htpp]
\centering
\includegraphics[scale=0.8]{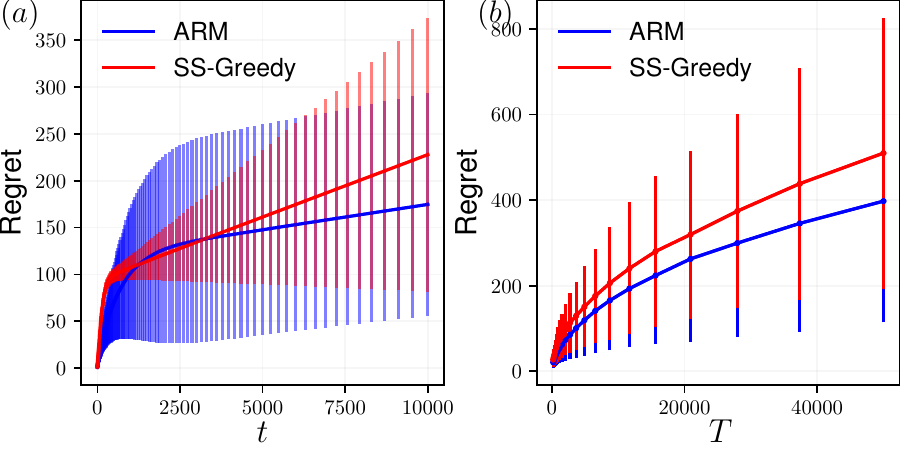}
\caption{Mean regret for the many-armed bandit problem with Bernoulli rewards and a uniform prior on $[0,1]$. This figure is identical to the one given in the main text but with standard deviation shown with error bars. \textbf{(a)} The regret growth of our algorithm (\algonamesemany), in blue, is observed until horizon $\hori$, i.e the stopping time, is reached $\hori=10000$ ($K=5000 \gg \sqrt{\hori}$ and $c=1$). The regret is averaged over $2000$ runs and standard error are indicated  (see Supplementary Material \cref{supp:Secnumericalexpe,supp:Secotheralgo} for details on the numerical settings). 
Its performance is compared to the ss-greedy algorithm, in red. The final slopes of both algorithms differs indicating that the selected arm for most of the exploitation is less efficient than the one found for our algorithm. Additionally, our algorithm presents a slower initial slope because exploration and exploitation are already mixed. \textbf{(b)} Mean regret performance when at the stopping time (when the horizon is reached),  for distinct games with varying horizon length ($K=\hori \gg \sqrt{\hori}$ and $c=1$). The regret is averaged over $2000$ runs.
}
\label{suppfig:3pres}
\end{figure}

\end{widetext}
\end{document}